\documentclass[10pt,twocolumn,letterpaper]{article}

\usepackage[pagenumbers]{cvpr} % To force page numbers, e.g. for an arXiv version

\definecolor{cvprblue}{rgb}{0.21,0.49,0.74}
\usepackage[pagebackref,breaklinks,colorlinks,allcolors=cvprblue]{hyperref}

\usepackage{algorithm}
\usepackage{listings}

\def\paperID{} % *** Enter the Paper ID here
\def\confName{CVPR}
\def\confYear{2025}

\newcommand{\authorskip}{\hspace{4.8mm}}

\definecolor{pink}{rgb}{1.0, 0.35, 0.35}
\newcommand{\revised}[1]{\textcolor{black}{#1}}

\usepackage{multirow}
\usepackage{colortbl}
\definecolor{lighterpurple}{rgb}{0.8, 0.6, 0.8}
\definecolor{lightorange}{rgb}{0.988, 0.824, 0.6}

\title{ObjCtrl-2.5D: Training-free Object Control with Camera Poses \\
\vspace*{10pt}
\small Project page:  \url{https://wzhouxiff.github.io/projects/ObjCtrl-2.5D/}
\vspace*{-10pt}
}

\author{
Zhouxia Wang \hspace{2.5mm}
\authorskip Yushi Lan\hspace{2.5mm}
\authorskip Shangchen Zhou  \hspace{2.5mm}
\authorskip Chen Change Loy \hspace{2.5mm}
\\ [2mm]
{
\fontsize{12pt}{12pt}\selectfont
S-Lab, Nanyang Technological University
}
}

\begin{document}

\twocolumn[{%
\renewcommand\twocolumn[1][]{#1}%
\maketitle
\begin{figure}[H]
\vspace{-1cm}
\hsize=\textwidth
\centering
% \fbox{\rule{0pt}{4in} \rule{0.9\linewidth}{0pt}}
\includegraphics[width=6.8 in]{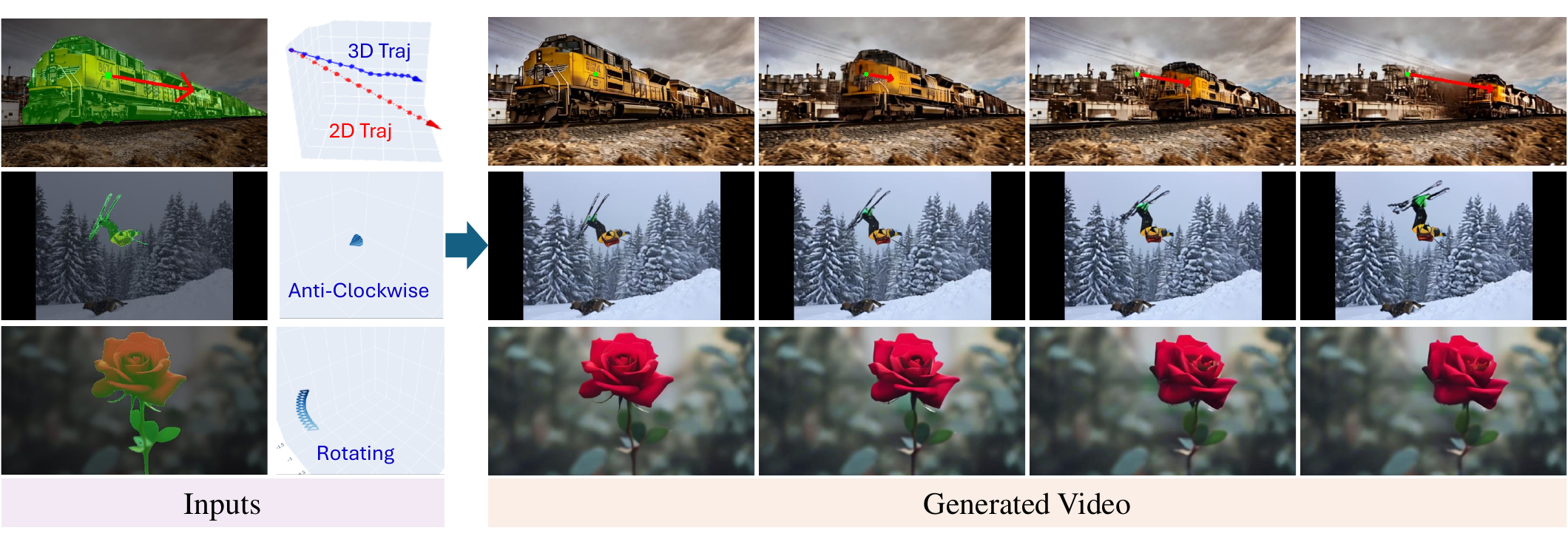}
\vspace{-0.3cm}
% \caption{ObjCtrl-2.5D enables versatile object motion control for image-to-video generation. It accepts 2D trajectories, 3D trajectories, or camera poses as control guidance (all transformed to camera poses) and achieves precise motion control by utilizing an existing camera motion control module \textbf{without additional training}. \textit{Unlike existing methods based on 2D trajectories, ObjCtrl-2.5D supports complex motion control beyond planar movement, such as object rotation, as demonstrated in the last row}. \textbf{We strongly recommend viewing the \href{https://wzhouxiff.github.io/projects/ObjCtrl-2.5D/}{project page} for dynamic results.}}
\caption{\textbf{Control Results of ObjCtrl-2.5D}. ObjCtrl-2.5D enables versatile object motion control for image-to-video generation. It accepts 2D trajectories (transformed to 3D), or camera poses as control guidance (all transformed to camera poses) and achieves precise motion control by utilizing an existing camera motion control module \textbf{without additional training}. Unlike existing methods based on 2D trajectories, ObjCtrl-2.5D supports complex motion control beyond planar movement, such as object rotation in the last row. \textbf{We strongly encourage consulting our \href{https://wzhouxiff.github.io/projects/ObjCtrl-2.5D/}{project page} for dynamic results, as they cannot be effectively represented through still images.}}
\label{fig:teaser} 
\end{figure}
}]

\maketitle
\begin{abstract}

This study aims to achieve more precise and versatile object control in image-to-video (I2V) generation. Current methods typically represent the spatial movement of target objects with 2D trajectories, which often fail to capture user intention and frequently produce unnatural results.
To enhance control, we present ObjCtrl-2.5D, a training-free object control approach that uses a 3D trajectory, extended from a 2D trajectory with depth information, as a control signal. By modeling object movement as camera movement, ObjCtrl-2.5D represents the 3D trajectory as a sequence of camera poses, enabling object motion control using an existing camera motion control I2V generation model (CMC-I2V) without training.
To adapt the CMC-I2V model originally designed for global motion control to handle local object motion, we introduce a module to isolate the target object from the background, enabling independent local control. In addition, we devise an effective way to achieve more accurate object control by sharing low-frequency warped latent within the object's region across frames. 
Extensive experiments demonstrate that ObjCtrl-2.5D significantly improves object control accuracy compared to training-free methods and offers more diverse control capabilities than training-based approaches using 2D trajectories, enabling complex effects like object rotation.

% \keywords{Object Motion Control; Image-to-Video Generation; Diffusion Model; Training-free}

\end{abstract}    
\section{Introduction}
\label{sec:intro}

\begin{figure*}[t]
    \centering
    \includegraphics[width=0.95\linewidth]{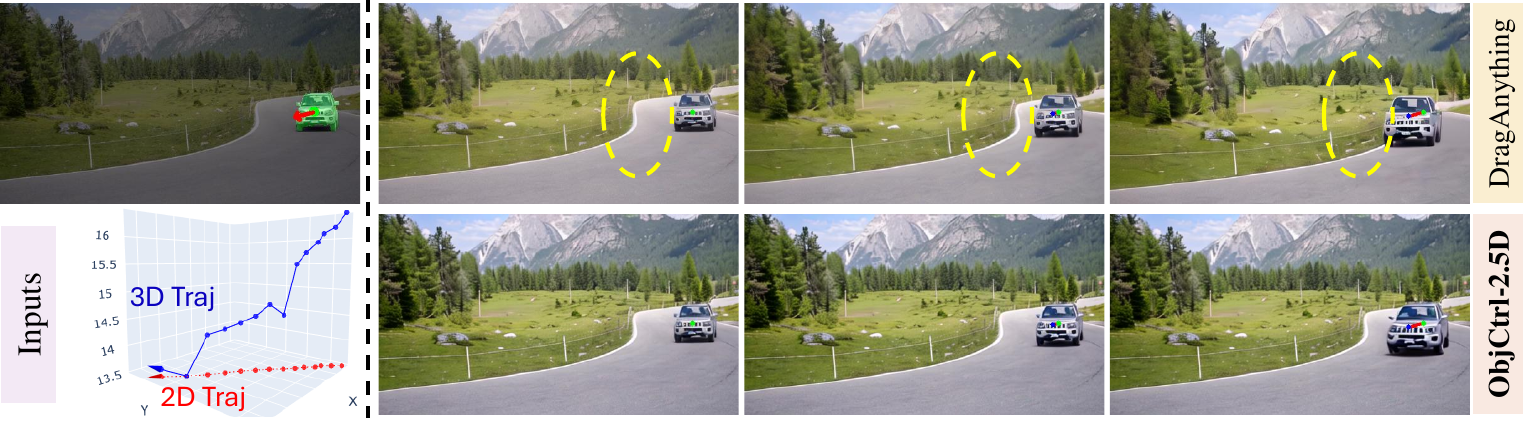}
    \caption{\textbf{Object control results using 2D and 3D trajectories.} On the left, the \textcolor{red}{red} line represents the 2D trajectory, the \textcolor{blue}{blue} line indicates the 3D trajectory extracted from real-world video in DAVIS~\cite{davis}, and the \textcolor{green}{green} point marks the starting point of the trajectory. The training-based method DragAnything~\cite{draganything}, which controls objects using a 2D trajectory, closely follows the specified path; however, it results in the car appearing to move horizontally toward the grass, which is atypical in real-world settings. By incorporating depth information from a 3D trajectory, our proposed method generates videos that not only follow the spatial trajectory but also achieve more realistic movement. 
    }
    \label{fig:2d_3d_traj}
    \vspace{-4mm}
\end{figure*}

Video generation seeks to produce high-quality videos from either a given text prompt (T2V generation) or a conditional image (I2V generation) and recently, numerous effective diffusion-based video generation models have emerged~\cite{VDM, LVDM, pixeldance, modelscope, videocrafter1, videocrafter2, dynamicrafter, svd, sora, open-sora, latte, cogvideo, cogvideox, lumiere}. The advancement of these models has spurred interest in developing more controllable generation, particularly for controlling the movement of objects within the generated video.

% Why 3D
Most existing methods control objects using two-dimensional (2D) representations, such as bounding boxes~\cite{boximator,peekaboo,trailblazer,freetraj,directavideo} and trajectories composed of discrete points~\cite{dragnvwa,motionctrl,imageconductor,draganything}. These 2D guides specify only the spatial position of the moving object, while real-world objects move within a three-dimensional (3D) space. The lack of 3D information often results in unnatural video outputs, as illustrated in Fig.~\ref{fig:2d_3d_traj}. The first row presents a result generated by DragAnything~\cite{draganything}, a training-based object control method that relies on 2D trajectories. Although the car relatively accurately follows the provided 2D trajectory, its movement is almost entirely horizontal toward the grass, which is unrealistic. In contrast, the second row shows a 3D trajectory extracted from the real-world video. It indicates that the car moves not only toward the lower-left direction but also approaches the camera, with depth decreasing from 16.5 to 13.5, which can explicitly guide the car to move along the road, rather than veering off into the grass.

To more effectively leverage such valuable depth information, we propose ObjCtrl-2.5D\footnote{Our approach is termed 2.5D because, while combining a 2D trajectory with depth information produces a 3D trajectory that enables more realistic and controlled simulations of object movement in 3D space, it does not capture all aspects of 3D geometry.} to significantly enhance object motion control accuracy in T2V generation by explicitly leveraging 3D trajectories derived from 2D trajectories and scene depth information. Inspired by the effectiveness of camera motion control using camera poses in vision generation, such as MotionCtrl~\cite{motionctrl} and CameraCtrl~\cite{cameractrl}, we propose to model object movement with camera poses, which allows us to fully utilize the existing Camera Motion Control T2V (CMC-T2V) model for object motion control without additional training.

Specifically, we first extend the 2D trajectory into 3D by incorporating depth information extracted from the conditional image. The resulting 3D trajectory is then converted into a sequence of camera poses through triangulation~\cite{hartley2003multiple}. To adapt existing global camera motion control models~\cite{motionctrl,cameractrl} for localized object motion control, we propose a Layer Control Module (LCM). This module disentangles the target object from the background, enabling independent motion control for the foreground object and the surrounding scene, without the need for training.
Additionally, we propose a Shared Warping Latent (SWL) to further improve object control accuracy by sharing low-frequency warping latents within the object’s area in each frame, establishing an initial object movement that significantly influences the subsequent generation process. Leveraging 3D information and a carefully designed object control model based on camera poses, ObjCtrl-2.5D achieves a significant improvement in control accuracy compared to previous training-free object control methods~\cite{peekaboo, trailblazer, freetraj}. Furthermore, as ObjCtrl-2.5D can accept custom camera pose sequences, it allows for more complex object motion control, such as object rotation, as illustrated in Fig.~\ref{fig:teaser}.

In conclusion, this work makes the following main contributions: 1) ObjCtrl-2.5D extends 2D trajectories to 3D using depth information and represents these 3D signals with camera poses, achieving training-free object motion control with higher accuracy. 2) ObjCtrl-2.5D introduces a Layer Control Module and Shared Warping Latent, adapting the camera motion control module for effective object motion control and significantly enhancing object control performance. 3) ObjCtrl-2.5D achieves more complex and diverse object control capabilities compared to previous 2D-based methods.

In the remainder of this paper, we first review the existing literature on video generation and diffusion-based object motion control methods in Section~\ref{sec:related}. Section~\ref{sec:methodology} presents the details of our proposed ObjCtrl-2.5D framework. In Section~\ref{sec:exp}, we conduct extensive experiments, including both quantitative and qualitative analyses, to validate the effectiveness of our approach. Finally, we conclude the paper and summarize our contributions in Section~\ref{sec:conclusion}.
\section{Related Work}
\label{sec:related}
\label{sec:method}

\noindent\textbf{Video Generation.}
With the rising interest in content generation, video generation has become a prominent research area, producing a wealth of impactful work based on generative adversarial networks (GAN)~\cite{VGAN, FTGAN, MoCoGAN, TGAN, TGANv2, DVD-GAN, G3AN} and diffusion models (DM)~\cite{VDM, LVDM, pixeldance, modelscope, videocrafter1, videocrafter2, dynamicrafter, svd, sora, open-sora, latte, cogvideo, cogvideox, lumiere}. 
Compared to GAN-based methods, diffusion models offer substantial advantages. To maximize the use of high-quality image datasets, most DM-based video generation models are derived from robust image-generation models, incorporating temporal modules and fine-tuning on video datasets. Notable examples include VDM~\cite{VDM}, which is based on a pixel space diffusion model, and LVDM~\cite{LVDM}, which extends a latent diffusion model. Numerous models follow a similar framework, such as Align-Your-Latents~\cite{blattmann2023align}, AnimateDiff~\cite{guo2023animatediff}, the VideoCrafter series~\cite{videocrafter1, videocrafter2, dynamicrafter}, and SVD~\cite{svd}, among others. 
Furthermore, recent studies reveal that diffusion models based on transformers (DiT)~\cite{sora, open-sora, latte, cogvideo, cogvideox} enhance both generation quality and scalability in video generation by replacing the conventional U-Net~\cite{unet} backbone with a transformer architecture. 
%Given the effectiveness of DiT-based methods in video generation, 
This study adopts the U-Net-based diffusion model SVD~\cite{svd}, as it is relatively mature in video generation and includes various extensions, such as control modules~\cite{cameractrl}, which are valuable for exploring object control in this work. Besides, as an image-to-video generation model, SVD can tie the object and trajectories easily by drawing trajectory on the given conditional image.

\begin{figure*}[t]
    \centering
    \includegraphics[width=0.98\linewidth]{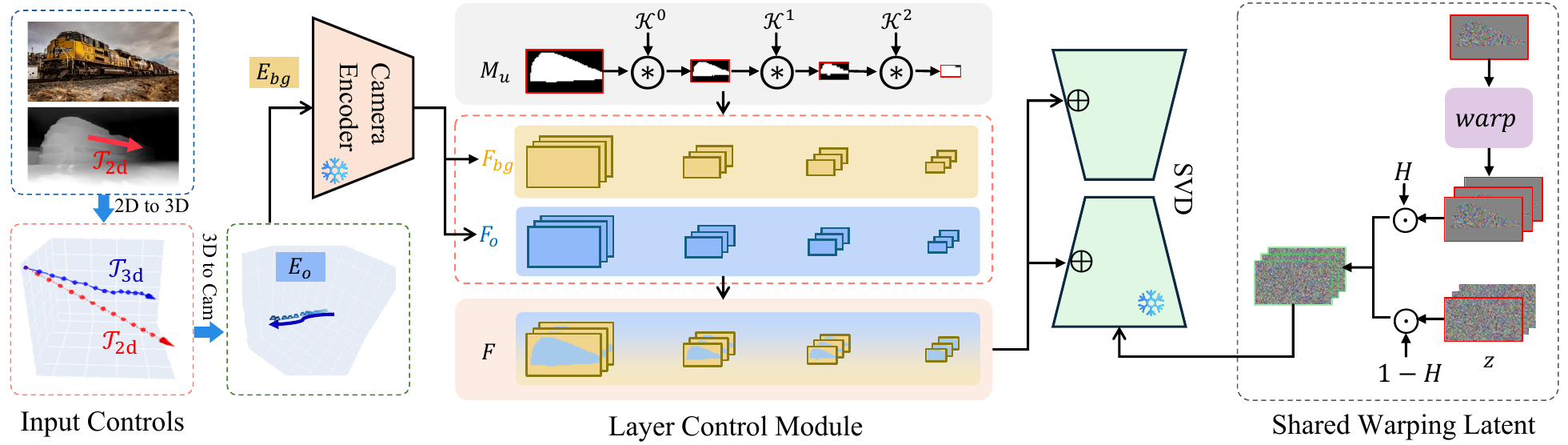}
    \vspace{2mm}
    \caption{\textbf{Framework of ObjCtrl-2.5D.} ObjCtrl-2.5D first extends the provided 2D trajectory $\mathcal{T}_{2d}$ to a 3D trajectory $\mathcal{T}_{3d}$ using depth information from the conditioning image. This 3D trajectory is then transformed into a camera pose $\mathbf{E_o}$ via triangulation~\cite{hartley2003multiple}. To achieve object motion control within a frozen camera motion control module, ObjCtrl-2.5D integrates a Layer Control Module (LCM) that separates the object and background with distinct camera poses ($\mathbf{E_o}$ and $\mathbf{E_{bg}}$). After extracting camera pose features via a Camera Encoder, LCM spatially combines these features using a series of scale-wise masks. Additionally, ObjCtrl-2.5D introduces a Shared Warping Latent (SWL) technique, implemented with a 3D low-pass filter $H$, to enhance control by sharing low-frequency initialized noise across frames within the warped areas of the object.}
    \label{fig:framework}
\end{figure*}

\noindent\textbf{Object Motion Control in Diffusion Video Models.} 
Advances in basic video generation have improved developments in video customization, including motion control for both camera and object movement. Although previous works, such as Tune-A-Video~\cite{tav}, MotionDirector~\cite{motiondirector}, LAMP~\cite{lamp}, VideoComposer~\cite{videocomposer}, and Control-A-Video~\cite{controlavideo}, enable motion learning from specific reference videos or guided motion generation through depth maps, sketches, or motion vectors derived from reference videos, these approaches often lack user-friendliness.  
Given their flexibility and interactivity, trajectory~\cite{chen2023motion,dragnvwa,motionctrl,draganything,imageconductor,dragdiffusion,animateanything,dragondiffusion,drag,revideo} and bounding box-based~\cite{peekaboo,trailblazer,boximator,directavideo,freetraj} methods have become popular in video motion control, generally classified as either training-based or training-free approaches. Training-based methods, including DragNUWA~\cite{dragnvwa}, DragAnything~\cite{draganything}, and ImageConductor~\cite{imageconductor}, utilize trajectories to control both camera and object motion, while Boximator~\cite{boximator} achieves control using bounding boxes. MotionCtrl~\cite{motionctrl}, by contrast, independently manages camera and object movements with separate camera and trajectory controls. Although effective, these methods demand significant computational resources for data curation and model training.
Alternatively, training-free methods, SG-I2V~\cite{SG-I2V} and ~\cite{xiao2024video} required per-sample optimization, and Direct-A-Video~\cite{directavideo}, PEEKABOO~\cite{peekaboo}, TrailBlazer~\cite{trailblazer}, and FreeTraj~\cite{freetraj}, enable object motion control by adjusting attention weights and initial noise according to specified trajectories and object bounding boxes. Although efficient and less computationally demanding, these methods are limited to 2D spatial object movements and can only coarsely constrain generated models within the given bounding boxes, which limits accuracy and the ability to model diverse movements. Although several concurrent works~\cite{levitor,PerceptionAsControl,DaS} also utilize 3D information for object motion control, they rely on carefully curated datasets for supervised training.

In contrast, ObjCtrl-2.5D is a training-free approach that achieves accurate and versatile object motion control in image-to-video (I2V) generation by carefully adapting existing camera motion control modules for object-level manipulation.
\section{Methodology}
\label{sec:methodology}

\subsection{Preliminaries}
Since ObjCtrl-2.5D is built on Stable Video Diffusion (SVD)~\cite{svd} and CameraCtrl~\cite{cameractrl}, we first provide a brief description of these two approaches before delving into our proposed method.

\noindent\textbf{Stable Video Diffusion (SVD)}. We adopt SVD~\cite{svd}, a publicly available and commonly used I2V diffusion model, as the basic model for our generation. SVD takes a conditional image $\mathbf{I_c}$ as input and generates a video with $N$ frames $\{\mathbf{F}^0, \mathbf{F}^1, \dots, \mathbf{F}^{N-1}\}$ using a conditional 3D U-Net~\cite{unet} integrated with a latent denoising diffusion process~\cite{ldm}.

\noindent\textbf{CameraCtrl.} Considering that object motion reflects the changes in spatial location across frames, we adopt CameraCtrl~\cite{cameractrl}, a model that spatially represents camera poses using Plücker embeddings~\cite{plucker}, as the basis for our object motion control.
Generally, camera poses include intrinsic parameters, denoted $\mathbf{K} = [[f_x, 0, c_x],[0, f_y, c_y],[0, 0, 1]]$, and extrinsic parameters $\mathbf{E}=[\mathbf{R|t}]$, where $\mathbf{R} \in \mathbb{R}^{3 \times 3}$ represents camera rotation and $\mathbf{t} \in \mathbb{R}^{3 \times 1}$ represents translation. Plücker embeddings enhance this representation by defining camera poses spatially as $\mathbf{p}_{x,y}=(\mathbf{o} \times \mathbf{d}_{x,y}, \mathbf{d}_{x,y}) \in \mathbb{R}^6$, where $(x,y)$ indicates a position in image coordinates, $\mathbf{o} \in \mathbb{R}^3$ is equal to $\mathbf{t}$ and represents the camera center in world coordinates, and $\mathbf{d}_{x,y} \in \mathbb{R}^3$ is the direction vector from the camera center to pixel $(x,y)$ in world coordinates. Specifically, 
\begin{equation}
    \mathbf{d}_{x,y} = \mathbf{RK^{-1}}[x, y, 1]^T + \mathbf{t}. 
\end{equation}
CameraCtrl extracts multi-scale camera motion information from the Plücker embeddings $\mathbf{P} \in \mathbb{R}^{N\times 6 \times H \times W}$, where $N$, $H$, and $W$ represent the length, height, and width of the generated video, respectively, using a camera encoder. This camera motion information is then integrated into SVD, enabling global camera motion control.

\subsection{ObjCtrl-2.5D}
ObjCtrl-2.5D is a training-free model for object motion control, distinguishing itself from previous 2D-based approaches~\cite{peekaboo,freetraj,dragnvwa,draganything} using 3D trajectories, which are attained by extending 2D trajectories with depth information. These 3D trajectories serve as control signals and are expressed as camera poses, allowing ObjCtrl-2.5D to leverage existing camera motion control models like CameraCtrl~\cite{cameractrl} for object motion control without additional training. 

Specifically, we first extend a 2D trajectory to 3D with depth from a conditional image. Subsequently, the 3D trajectory is modeled as a sequence of camera poses using triangulation~\cite{hartley2003multiple} (Section~\ref{sec:3d_to_cam}). To adapt global motion methods, such as CameraCtrl, to local motion control, we introduce a Layer Control Module (LCM) that isolates the target object from the background, allowing for independent local manipulation (Section~\ref{sec:LCM}). Additionally, Shared Warped Latents (SWL) is proposed to improve object control accuracy by sharing low-frequency warped latent information across the object area in each frame (Section~\ref{sec:SWL}).

\subsubsection{2D Trajectory to 3D to Camera Poses}
% \subsubsection{2D Trajectory $\rightarrow$ 3D $\rightarrow$ Camera Poses}
\label{sec:3d_to_cam}

% \text{}
% \\
\noindent\textbf{2D Trajectory to 3D.} The 2D trajectory is represented as $\mathcal{T}_{2d} = \{(x^0, y^0), (x^1, y^1), \dots, (x^{N-1}, y^{N-1})\}$, where $i \in [0, N-1]$ and $N$ is the number of frames. This trajectory is extended to 3D as $\mathcal{T}_{3d} = \{p^0, p^1, \dots, p^{N-1}\}$, with points $p^i = (x^i, y^i, d^i)$. $d^i$ is the depth value of $\mathbf{D_c}$ at the coordinate $(x^i, y^i)$, where the depth map $\mathbf{D_c}$ is extracted from the conditional image $\mathbf{I_c}$.
% using ZoeDepth~\cite{zoedepth}.

% To maintain smooth transitions, any abrupt depth changes between neighboring trajectory points are normalized. Additional details are provided in the supplementary materials.

\begin{figure}[t]
    \centering
    \includegraphics[width=0.86\linewidth]{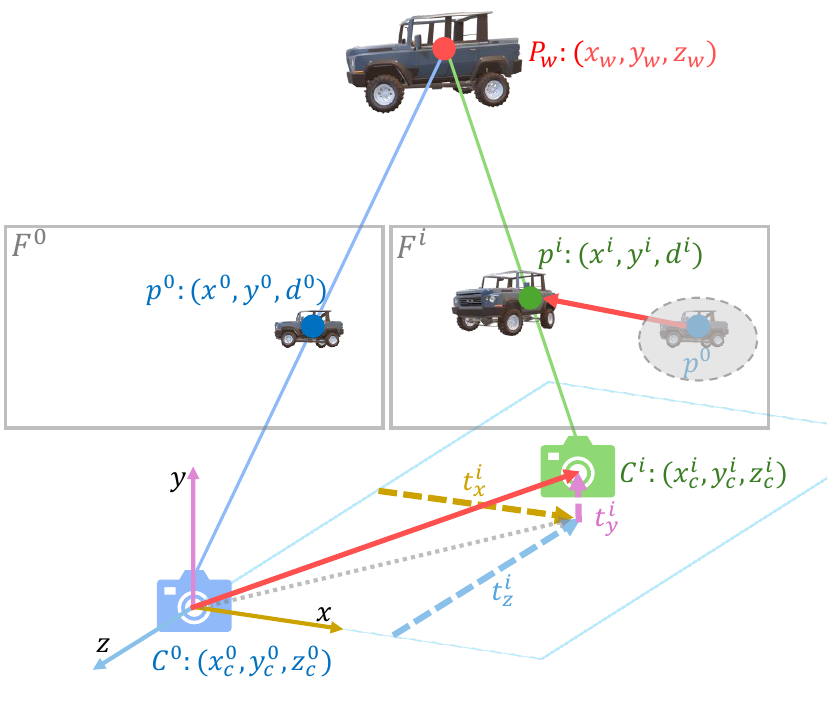}
    \vspace{-0.3cm}
    \caption{\textbf{3D Trajectory to Camera Poses.} We model the object movement in a video, indicated by a 3D trajectory, as the camera's location translation in 3D space. Details refer to Sec.~\ref{sec:3d_to_cam} and Algorithm.~\ref{alg:3d_to_cam}.
    }
    \label{fig:3d_to_cam}
    \vspace{-4mm}
\end{figure}

\noindent\textbf{3D Trajectory to Camera Poses.} 
In this work, we transform the 3D trajectory to camera poses with triangulation~\cite{hartley2003multiple}.
As illustrated in Fig.~\ref{fig:3d_to_cam}, the object’s movement from $p^0$ to $p^i$ between frames $\mathbf{F}^0$ and $\mathbf{F}^i$ is modeled as a corresponding camera movement from $\mathbf{C}^0$ to $\mathbf{C}^i$, with all trajectory points mapped to the same point $\mathbf{P}_w = (x_w, y_w, z_w)$ in world coordinates. Since user-provided trajectories are often sparse, making it difficult to recover extrinsic parameters with both rotation $\mathbf{R}$ and translation $\mathbf{t}$, we simplify by modeling the 3D trajectory as camera translation only, omitting rotation. Thus, $\mathbf{R}$ is set as an identity matrix $\mathbf{I}$ for all camera poses, allowing us to represent the 3D trajectory with camera movement by solving for $\mathbf{t}^i = [t^i_x, t^i_y, t^i_z]$ using triangulation~\cite{hartley2003multiple}.

Specifically, we first project each point $p^i = (x^i, y^i, d^i)$ from image space into camera coordinates $\mathbf{C}^i = (x^i_c, y^i_c, z^i_c)$ using the camera intrinsic matrix $\mathbf{K} = [[f_x, 0, c_x],[0, f_y, c_y],[0, 0, 1]]$:
\begin{equation}
    x^i_c = z^i_c (x^i - c_x) / f_x; \; y^i_c = z^i_c (y^i - c_y) / f_y; \; z^i_c = d^i.
    \label{eq:pixel2camera}
\end{equation}
% Following previous works~\cite{spatracker,som}, $\mathbf{K}$ can be roughly estimated based on the spatial dimensions of the generated video or estimated with existing methods, such as UniDepth~\cite{unidepth}. 
%
Then, we compute $\mathbf{P}_w = (x_w, y_w, z_w)$ in world space with world-to-camera transformation, \ie, $\mathbf{C}^i = [\mathbf{I}|\mathbf{t}^i][x_w, y_w, z_w, 1]^T$, attained:
\begin{equation}
    x_w = x^i_c - t^i_x; \quad y_w = y^i_c - t^i_y; \quad z_w = z^i_c - t^i_z.
    \label{eq:c2w}
\end{equation}
Following DUSt3R~\cite{dust3r}, we set the first frame $\mathbf{F}^0$ as the canonical camera space, \ie, $\mathbf{t}^0 = [0, 0, 0]$, and the subsequent frames are expressed in the same coordinate space as $\mathbf{F}^0$. Thus, $\mathbf{P}_w = (x^0_c, y^0_c, z^0_c)$ and:
\begin{equation}
    t^i_x = x^i_c - x^0_c; \; t^i_y = y^i_c - y^0_c; \; t^i_z = z^i_c - z^0_c.
\end{equation}
Essentially, we provide a Python implementation code in Algorithm~\ref{alg:3d_to_cam}.

Note that while ObjCtrl-2.5D models 3D trajectories as camera poses without incorporating rotation by default, the translation-based camera movement alone already surpasses traditional 2D trajectories, resulting in more accurate and natural video generation (see Fig.~\ref{fig:2d_3d_traj}). Furthermore, ObjCtrl-2.5D supports user-defined camera poses with rotational components, thereby enabling rotational object motion control (see Fig.~\ref{fig:teaser}, Fig.~\ref{fig:camera_pose}, and Fig.~\ref{fig:more_results}).

\subsubsection{Layer Control Module}
\label{sec:LCM}

To adapt CameraCtrl~\cite{cameractrl}, originally designed for global motion control, to object-specific motion, we introduce  Layer Control Module (LCM). This module separates the conditional image $\mathbf{I_c}$ into foreground and background layers using an object mask $\mathbf{M}_c$.
% generated via instance segmentation, such as SAM~\cite{sam,sam2}
The foreground layer is controlled by object-specific camera poses $\mathbf{E_o}$, derived from 3D trajectories outlined in Sec.~\ref{sec:3d_to_cam}, while the background layer is guided by background-specific poses $\mathbf{E_{bg}}$. 
% These background poses can be customized, with options like $[\mathbf{I}|\mathbf{0}]$ allowing for a static background.

To extract camera features, $\mathbf{E_o}$ and $\mathbf{E_{bg}}$ are fed into the Camera Encoder in~\cite{cameractrl}, yielding $\mathbf{F_o} = \{f_o^0, f_o^1, \dots, f_o^{S-1}\}$ and $\mathbf{F_{bg}} = \{f_{bg}^0, f_{bg}^1, \dots, f_{bg}^{S-1}\}$, as UNet~\cite{unet} in SVD~\cite{svd} contains feature maps with $S$ scales.
% These features are then fused with mask $\mathbf{M_o}$, which indicates the dominated area of $\mathbf{E_o}$, while $(1-\mathbf{M_o})$ indicates the dominated area of $\mathbf{E_{bg}}$.
%
To ensure $\mathbf{E_o}$ comprehensively covers the areas of the moving object across all the frames, we first attain the frame-wise object area $\mathbf{M_w} = \{ m_w^0, m_w^1, \dots, m_w^{N-1} \}$ from $\mathbf{M}_c$ using a geometric warping function $\mathrm{warp}(\cdot)$~\cite{hartley2003multiple,luciddreamer,genwarp}, where:
\begin{equation}
    m_w^i = \mathrm{warp}(\mathbf{M}^0; \mathbf{D_c}, \mathbf{E_o}^{0}, \mathbf{E_o}^{i}, \mathbf{K}), \quad i \in [0, N-1].
    \label{eq:warp_mask}
\end{equation}
$\mathbf{D_c}$ is the depth, $\mathbf{E_o}^{i}$ is the object’s camera pose for frame $i$, and $\mathbf{K}$ represents the intrinsic matrix. The union of these masks, $\mathbf{M_u} = \bigcup_{i=0}^{N-1} m_w^i$, defines the complete object area dominated by $\mathbf{E_o}$.

To prevent $\mathbf{M_u}$ from losing effectiveness during smaller-scale feature fusion, particularly for smaller target objects, we introduce a scale-wise mask strategy, which progressively dilates $\mathbf{M_u}$ at each scale using kernel $\mathcal{K}$. This process generates a set of dilated masks $\mathbf{M_o}=\{m_o^0, m_o^1,\dots,m_o^{S-1}\}$, where
\begin{equation}
    m_o^{s} = m_o^{s-1} \ast \mathcal{K}^{s-1}, \quad s \in [0, S-1], \quad m_o^{-1} = \mathbf{M_u}.
\end{equation}
Then fused feature $\mathbf{F} = \{f^0, f^1, \dots, f^{S-1}\}$ is:
\begin{equation}
    f^s = f_o^s \odot \mathbf{m_o}^{s} + f_{bg}^s \odot (1 - \mathbf{m_o}^{s}), \quad s \in [0, S-1],
\end{equation}
which is scale-wisely injected into SVD to control object motion in the generated video.

%##################################################################################################
\begin{algorithm}[t]
\caption{\small Implementation of 3D Trajectory to Camera Poses.}
\label{alg:3d_to_cam}
%\vspace{-1.em}

\definecolor{codeblue}{rgb}{0.25,0.5,0.5}
\lstset{
  backgroundcolor=\color{white},
  basicstyle=\fontsize{7.2pt}{7.2pt}\ttfamily\selectfont,
  columns=fullflexible,
  breaklines=true,
  captionpos=b,
  commentstyle=\fontsize{7.2pt}{7.2pt}\color{codeblue},
  keywordstyle=\fontsize{7.2pt}{7.2pt},
%  frame=tb,
}
\begin{lstlisting}[language=python]

def Traj3D_to_CameraPoses(T3d, fx, fy, cx, cy):
    # Input:
    #     T3d: numpy.array, [N, 3], [frame_id, (x, y, d)]
    #     fx, fy, cx, cy: float, intrinsic paramters.
    # Output:
    #     t: [tx, ty, tz]
    
    zc = T3d[:, 2]
    xc = (T3d[:, 0] - cx) * zc / fx
    yc = (T3d[:, 1] - cy) * zc / fy

    xw, yw, zw = xc[0], yc[0], zc[0]
    tx, ty, tz = xc - xw, yc - yw, zc - zw
    
    return [tx, ty, tz]

\end{lstlisting}
% \vspace{-10mm}
\end{algorithm}
% \vspace{-4mm}
%##################################################################################################

\begin{figure*}[t]
    \centering
    \includegraphics[width=0.96\linewidth]{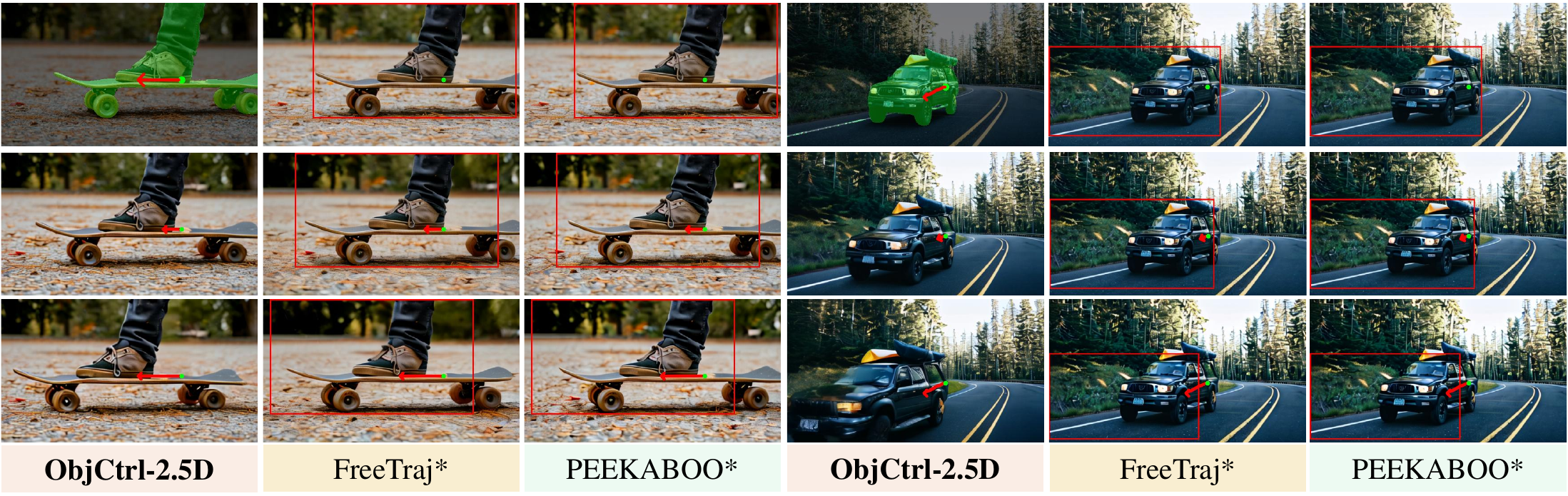}
    \caption{\textbf{Qualitative Comparison with Training-free Methods.} Compared to PEEKABOO~\cite{peekaboo} and FreeTraj~\cite{freetraj} that coarsely move the objects within the bounding boxes derived from the trajectory, our ObjCtrl-2.5D achieves higher trajectory alignment by extending the 2D trajectory to 3D and accurately transforming it into camera poses through triangulation~\cite{hartley2003multiple}.}
    \label{fig:comp_training_free}
    % \vspace{-3mm}
    \vspace{-0.2cm}
\end{figure*}

\begin{figure*}[h]
    \centering
    % \vspace{-3mm}
    \includegraphics[width=0.99\linewidth]{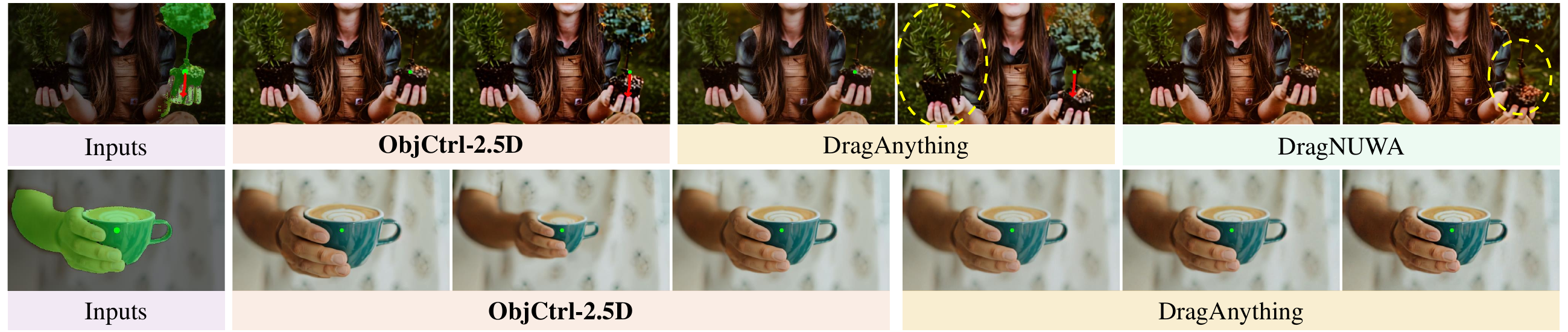}
    % \vspace{-3mm}
    \caption{\textbf{Qualitative Comparison with Training-based Methods.} Due to their training strategy, DragAnything~\cite{draganything} tends to apply global movement to objects (both potted plants shift downward, despite only the right plant being specified to move), and DragNUWA~\cite{dragnvwa} often moves only part of the target object. In contrast, our proposed ObjCtrl-2.5D achieves precise, targeted object control thanks to its Layer Control Module. Additionally, ObjCtrl-2.5D is capable of performing more versatile object control when given a trajectory with a fixed spatial position (the \textcolor{green}{green} point in the second sample), such as front-to-back-to-front movement, while DragAnything~\cite{draganything} generates a relatively static video.}
    \label{fig:comp_training}
\end{figure*}

\subsubsection{Shared Warping Latent}
\label{sec:SWL}

As a training-free approach, ObjCtrl-2.5D with LCM achieves good performance in object motion control compared to related methods. To further enhance control accuracy on challenging cases, such as generating uncommon object movements like a reversing boat (as shown in Fig.~\ref{fig:ablation_lcm}), we introduce frame-wise shared low-frequency latents~\cite{freetraj}, \ie, Shared Warping Latent (SWL). Unlike FreeTraj~\cite{freetraj}, which simply copies object latents, bounding with a box, from the first frame to all frames, we employ the geometric warping function $\mathrm{warp}(\cdot)$~\cite{hartley2003multiple,luciddreamer,genwarp} to warp shared latent across frames, enabling a more precise object moving control.

Similar to Eq.~\ref{eq:warp_mask}, given the initial noise $\mathbf{z}$ of all the frame, we create a sequence of warped noise maps, $\mathbf{z}_w = \{\mathbf{z}_w^0, \mathbf{z}_w^1, \dots, \mathbf{z}_w^{N-1}\}$ from $\mathbf{z}^0$, the first noise map in $\mathbf{z}$, as follows:
\begin{equation}
    \mathbf{z}_w^i = \mathrm{warp}(\mathbf{z}^0; \mathbf{D_c}, \mathbf{E_o}^{0}, \mathbf{E_o}^{i}, \mathbf{K}), \quad i \in [0, N-1].
    \label{eq:warp_latent}
\end{equation}
To ensure that only latents within the object regions are shared across frames while preserving randomness in the background, we apply warping masks $\mathbf{M_w}$ to the warped noise, blending them back into $\mathbf{z}$ to produce $\mathbf{z_L}$:
\begin{equation}
    \mathbf{z_L} = \mathbf{M_w} \odot \mathbf{z_w} + (1 - \mathbf{M_w}) \odot \mathbf{z}.
\end{equation}
To mitigate the quality decrease of the generated video, only low-frequency information from $\mathbf{z_L}$ is retained:
\begin{equation}
    \mathbf{\hat{z}} = \mathcal{F F T}_{3 D} (\mathbf{z_L}) \odot \mathcal{H} + \mathcal{F F T}_{3 D} (\mathbf{z}) \odot (1 - \mathcal{H}),
\end{equation}
where $\mathcal{F F T}_{3 D}$ denotes the 3D Fast Fourier Transform~\cite{cooley1965algorithm,wu2023freeinit}, $\mathcal{H}$ is a 3D low-pass filter, and $\mathbf{\hat{z}}$ serves as the noise at the $T_{th}$ step in SVD.
\section{Experiments}
\label{sec:exp}

%\subsection{Experiments Settings}
\subsection{Experimental Details}

\begin{table*}[th]
  \centering
  
  \caption{\textbf{Quantitative Comparisons on DAVIS~\cite{davis}}. \revised{ObjCtrl-2.5D, as a training-free approach, shows promising improvement in object motion control compared to prior training-free methods, as indicated by ObjMC scores. Although there remains room for improvement compared to training-based methods, ObjCtrl-2.5D offers more versatile object control, as demonstrated in Fig.~\ref{fig:teaser} and Fig.~\ref{fig:comp_training}. 
  % Notably, the generated quality is largely determined by the based model (SVD), resulting in minimal differences between object control models in terms of FID, FVD, Imaging Quality, and Aesthetic Quality.
  }
  }
  \vspace{-0.2cm}
  % \begin{tabular}{@{}lc@{}lc@{}lc@{}lc@{}}
  \resizebox{\linewidth}{!}{
  % \begin{tabular}{c|ccccccc|ccccccc}
  \begin{tabular}{c|ccccccc} %|ccccccc}
    \toprule
    % \multicolumn{2}{c|}{Method} 
    % & \multicolumn{7}{c}{DAVIS} \\ %& \multicolumn{7}{c}{ObjCtrl-Test}   \\
    & \multirow{2}{*}{\textbf{FID~$\downarrow$}} 
    & \multirow{2}{*}{\textbf{FVD~$\downarrow$}}
    & \multirow{2}{*}{\textbf{ObjMC~$\downarrow$}}
    & \textbf{Imaging~$\uparrow$}
    & \textbf{Aesthetic~$\uparrow$}
    & \textbf{Motion~$\uparrow$}
    & \textbf{Dynamic~$\uparrow$} 
    % & \multirow{2}{*}{\textbf{FID~$\downarrow$}} 
    % & \multirow{2}{*}{\textbf{FVD~$\downarrow$}}
    % & \multirow{2}{*}{\textbf{ObjMC~$\downarrow$}}
    % & \textbf{Imaging~$\uparrow$}
    % & \textbf{Aesthetic~$\uparrow$}
    % & \textbf{Motion~$\uparrow$}
    % & \textbf{Dynamic~$\uparrow$}
    \\
   Methods 
   &  
   &  
   & 
   & \textbf{Quality}
   & \textbf{Quality}
   & \textbf{Smoothness}
   & \textbf{Degree}
   % &  
   % &  
   % & 
   % & \textbf{Quality}
   % & \textbf{Quality}
   % & \textbf{Smoothness}
   % & \textbf{Degree}
   \\
   \midrule
   \cellcolor{lighterpurple}\textbf{Training-Based Methods} &&&&&&& \\%&&&&&&& \\
   DragNUWA~\cite{dragnvwa} 
   & 62.36 & 11.68 & 37.57 
   & 0.6028 & 0.5033 & 0.9840 & 0.3441
   % & 235.94 & 27.45 & 58.80 
   % & 0.6201 & 0.5934 & 0.9874 & 0.3846
   \\
   DragAnything~\cite{draganything} 
   & 59.81 & 11.05 & 46.10 
   & 0.5729 & 0.4802 & 0.9786 & 0.4839
   % & 227.72 & 26.93 & 60.81
   % & 0.5821 & 0.5620 & 0.9843 & 0.7436
   \\
   \midrule
   \cellcolor{lighterpurple}\textbf{Training-free Methods} &&&&&&& \\ %&&&&&&& \\
   PEEKABOO$^*$~\cite{peekaboo} 
   & 62.43 & 11.97 & \cellcolor{lightorange}128.05 
   & 0.6195 & 0.5139  & 0.9830 & \cellcolor{lightorange}0.2472
   % & 250.68 & 27.54 & \cellcolor{lightorange}164.40 
   % & 0.6389 & 0.5909 & 0.9894 & \cellcolor{lightorange}0.1410
   \\
   FreeTraj$^*$~\cite{freetraj} 
   & 69.72 & 12.62 & \cellcolor{lightorange}125.30 
   & 0.6179 & 0.5111  & 0.9822 & \cellcolor{lightorange}0.3034
   % & 244.88 & 26.74 & \cellcolor{lightorange}158.39
   % & 0.6387 & 0.5913 & 0.9876 & \cellcolor{lightorange}0.2949
   \\
   % \midrule
   % \rowcolor{lighterpurple}
   \textbf{ObjCtrl-2.5D} 
   & 59.77 & 12.22 & \cellcolor{lightorange}\textbf{91.42} 
   & 0.6142 & 0.5087 & 0.9823 & \cellcolor{lightorange}\textbf{0.3483}
   % & 247.48 & 27.82 & \cellcolor{lightorange}\textbf{120.37}
   % & 0.61305 & 0.5825 & 0.9920 & \cellcolor{lightorange}\textbf{0.3461}
   \\
    \bottomrule
  \end{tabular}
  }
  % \vspace{-0.1cm}
  % \vspace{-20pt}
  \label{tab:sota_davis}
  % \vspace{-3mm}
\end{table*}

\begin{table*}[th]
  \centering
  % \vspace{-0.6cm}
  \caption{\textbf{Quantitative Comparisons on ObjCtrl-Test}. \revised{ObjCtrl-2.5D, as a training-free approach, shows promising improvement in object motion control compared to prior training-free methods, as indicated by ObjMC scores. Although there remains room for improvement compared to training-based methods, ObjCtrl-2.5D offers more versatile object control, as demonstrated in Fig.~\ref{fig:teaser} and Fig.~\ref{fig:comp_training}. 
  % Notably, the generated quality is largely determined by the based model (SVD), resulting in minimal differences between object control models in terms of FID, FVD, Imaging Quality, and Aesthetic Quality.
  }
  }
  % \begin{tabular}{@{}lc@{}lc@{}lc@{}lc@{}}
  \resizebox{\linewidth}{!}{
  % \begin{tabular}{c|ccccccc|ccccccc}
  \begin{tabular}{c|ccccccc} %|ccccccc}
    \toprule
    % \multicolumn{2}{c|}{Method} 
    % & \multicolumn{7}{c}{DAVIS} \\ %& \multicolumn{7}{c}{ObjCtrl-Test}   \\
    & \multirow{2}{*}{\textbf{FID~$\downarrow$}} 
    & \multirow{2}{*}{\textbf{FVD~$\downarrow$}}
    & \multirow{2}{*}{\textbf{ObjMC~$\downarrow$}}
    & \textbf{Imaging~$\uparrow$}
    & \textbf{Aesthetic~$\uparrow$}
    & \textbf{Motion~$\uparrow$}
    & \textbf{Dynamic~$\uparrow$} 
    % & \multirow{2}{*}{\textbf{FID~$\downarrow$}} 
    % & \multirow{2}{*}{\textbf{FVD~$\downarrow$}}
    % & \multirow{2}{*}{\textbf{ObjMC~$\downarrow$}}
    % & \textbf{Imaging~$\uparrow$}
    % & \textbf{Aesthetic~$\uparrow$}
    % & \textbf{Motion~$\uparrow$}
    % & \textbf{Dynamic~$\uparrow$}
    \\
   Methods 
   &  
   &  
   & 
   & \textbf{Quality}
   & \textbf{Quality}
   & \textbf{Smoothness}
   & \textbf{Degree}
   % &  
   % &  
   % & 
   % & \textbf{Quality}
   % & \textbf{Quality}
   % & \textbf{Smoothness}
   % & \textbf{Degree}
   \\
   \midrule
   \cellcolor{lighterpurple}\textbf{Training-Based Methods} &&&&&&& \\%&&&&&&& \\
   DragNUWA~\cite{dragnvwa} 
   % & 62.36 & 11.68 & 37.57 
   % & 0.6028 & 0.5033 & 0.9840 & 0.3441
   & 235.94 & 27.45 & 58.80 
   & 0.6201 & 0.5934 & 0.9874 & 0.3846
   \\
   DragAnything~\cite{draganything} 
   % & 59.81 & 11.05 & 46.10 
   % & 0.5729 & 0.4802 & 0.9786 & 0.4839
   & 227.72 & 26.93 & 60.81
   & 0.5821 & 0.5620 & 0.9843 & 0.7436
   \\
   \midrule
   \cellcolor{lighterpurple}\textbf{Training-free Methods} &&&&&&& \\ %&&&&&&& \\
   PEEKABOO$^*$~\cite{peekaboo} 
   % & 62.43 & 11.97 & \cellcolor{lightorange}128.05 
   % & 0.6195 & 0.5139  & 0.9830 & \cellcolor{lightorange}0.2472
   & 250.68 & 27.54 & \cellcolor{lightorange}164.40 
   & 0.6389 & 0.5909 & 0.9894 & \cellcolor{lightorange}0.1410
   \\
   FreeTraj$^*$~\cite{freetraj} 
   % & 69.72 & 12.62 & \cellcolor{lightorange}125.30 
   % & 0.6179 & 0.5111  & 0.9822 & \cellcolor{lightorange}0.3034
   & 244.88 & 26.74 & \cellcolor{lightorange}158.39
   & 0.6387 & 0.5913 & 0.9876 & \cellcolor{lightorange}0.2949
   \\
   % \midrule
   % \rowcolor{lighterpurple}
   \textbf{ObjCtrl-2.5D} 
   % & 59.77 & 12.22 & \cellcolor{lightorange}\textbf{91.42} 
   % & 0.6142 & 0.5087 & 0.9823 & \cellcolor{lightorange}\textbf{0.3483}
   & 247.48 & 27.82 & \cellcolor{lightorange}\textbf{120.37}
   & 0.61305 & 0.5825 & 0.9920 & \cellcolor{lightorange}\textbf{0.3461}
   \\
    \bottomrule
  \end{tabular}
  }
  % \vspace{-0.1cm}
  % \vspace{-20pt}
  \label{tab:sota_objctrl}
  % \vspace{-3mm}
\end{table*}

\noindent\textbf{Experimental Settings.}
We deploy ObjCtrl-2.5D on CameraCtrl-SVD~\cite{cameractrl}. ObjCtrl-2.5D supports various forms of object control input, including 2D trajectories, 3D trajectories, and complex camera poses, and generates videos with a resolution of $320 \times 576$ over 14 frames. In this paper, we adopt ZoeDepth~\cite{zoedepth} for depth map extraction and SAM2~\cite{sam2} for object masking. Note that the use of more advanced techniques may further improve performance.

\noindent\textbf{Evaluation Datasets.}
\textbf{(1) DAVIS:} To evaluate the effectiveness of ObjCtrl-2.5D on both 2D trajectories with depth and 3D trajectories, we extend the DAVIS dataset~\cite{davis} by generating 3D trajectories using SpatialTracker~\cite{spatracker}. The DAVIS dataset comprises 90 real-world videos with corresponding instance mask annotations. For each video, we use the first frame as the conditional image input for image-to-video (I2V) generation and randomly select one 3D trajectory within the instance mask as the guidance for object control.
\textbf{(2) ObjCtrl-Test:} \revised{Given that baseline I2V models often perform well on in-distribution trajectories extracted from real-world videos, we introduce a new synthetic test set, ObjCtrl-Test, to enable a more comprehensive evaluation. ObjCtrl-Test consists of 78 samples, each containing a high-quality image, an object mask specifying the target to be moved, and a corresponding 2D trajectory. Unlike DAVIS, which features motion patterns commonly observed in the real world, ObjCtrl-Test includes a diverse range of motions (such as cars moving backward), allowing for a more rigorous assessment of object motion control capabilities.}

\noindent\textbf{Evaluation Metrics.}
Following previous works~\cite{motionctrl,draganything}, we evaluate the generated video quality using the Fr\'{e}chet Inception Distance (FID)~\cite{fid} and Fr\'{e}chet Video Distance (FVD)~\cite{fvd}, taking the real videos in DAVIS~\cite{davis} as reference.
To assess object motion control precision, we use ObjMC~\cite{motionctrl}, which calculates the distance between target trajectories and the trajectories of generated videos, estimated using SpatialTracker~\cite{spatracker}. Lower ObjMC scores indicate better object control accuracy. 
\revised{Considering that FID and FVD can be biased, particularly when the reference set is small, we further incorporate evaluation metrics from VBench~\cite{vbench}, including Image Quality, Aesthetic Quality, Motion Smoothness, and Dynamic Degree.}
Additionally, we conduct a user study to provide a more comprehensive assessment of the effectiveness of ObjCtrl-2.5D.

\subsection{Comparison with State-of-the-art Methods}

To provide a thorough evaluation, we compare ObjCtrl-2.5D with both training-free and training-based methods. For training-free approaches, we use two recent methods: PEEKABOO~\cite{peekaboo} and FreeTraj~\cite{freetraj}. These methods, initially designed for I2V generation, incorporate adaptive attention mechanisms for object motion control. In adapting them for I2V generation, we omit manipulations on cross-attention since SVD~\cite{svd} utilizes a single embedding feature from the conditional image for cross-attention input. We denote these adapted versions as PEEKABOO$^*$ and FreeTraj$^*$. For training-based methods, we compare with DragNUWA~\cite{dragnvwa} and DragAnything~\cite{draganything}, both of which were trained with 2D trajectories and perform well under such conditions.

The quantitative results in Table~\ref{tab:sota_davis} and Table~\ref{tab:sota_objctrl} demonstrate that ObjCtrl-2.5D improves object motion control, as evidenced by the substantial reduction in the ObjMC score compared to other training-free methods. This improvement primarily stems from the fundamental differences in model design between ObjCtrl-2.5D and PEEKABOO$^*$ and FreeTraj$^*$. Both PEEKABOO$^*$ and FreeTraj$^*$ rely on 2D trajectories represented as a series of bounding boxes, as illustrated in Fig.~\ref{fig:comp_training_free}. This approach enables coarse object movement within the specified bounding boxes but lacks the precision of exact trajectory alignment. In contrast, ObjCtrl-2.5D achieves higher trajectory alignment by extending the 2D trajectory to 3D and accurately transforming it into camera poses through triangulation~\cite{hartley2003multiple}, yielding significantly better alignment with the given trajectory than PEEKABOO$^*$ and FreeTraj$^*$.

On the other hand, Table~\ref{tab:sota_davis} and Table~\ref{tab:sota_objctrl} indicate that ObjCtrl-2.5D remains room for improvement compared to training-based methods like DragNUWA~\cite{dragnvwa} and DragAnything~\cite{draganything}. These methods, trained on optical flow-based or tracker-derived trajectories, are inherently skilled at closely following specified trajectories, leading to high ObjMC performance. However, their design often results in moving the entire scene rather than isolating the target object's motion. This limitation is visible in DragAnything~\cite{draganything} in the first row of Fig.~\ref{fig:comp_training}, where both potted plants shift downward, despite only the right plant being specified to move. Moreover, in this example, DragNUWA~\cite{dragnvwa} fails to move the entire right-side plant, likely due to a lack of semantic awareness.
In contrast, ObjCtrl-2.5D achieves targeted object control advanced from the proposed Layer Control Module, which restricts the camera poses derived from the given trajectory to areas around the target object, minimally affecting the background. As demonstrated in the second row of Fig.~\ref{fig:comp_training}, when given a trajectory with a fixed spatial position, ObjCtrl-2.5D can perform front-to-back-to-front object movement by leveraging depth information (indicating an increase and subsequent decrease in depth). Meanwhile, DragAnything~\cite{draganything} tends to maintain object static in the generated video under similar conditions.

\revised{Regarding the generated quality, we observe that it is largely determined by the base model (SVD~\cite{svd}), resulting in minimal differences between object control methods in terms of FID, FVD, Imaging Quality, and Aesthetic Quality as in Table~\ref{tab:sota_davis} and Table~\ref{tab:sota_objctrl}}.

\begin{figure}[t]
    \centering
    \includegraphics[width=1.0\linewidth]{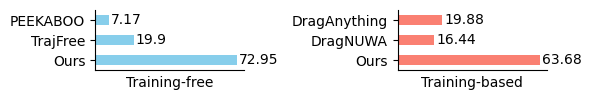}
    \vspace{-0.2cm}
    \caption{\textbf{User Study.} The majority of participants preferred the results obtained with ObjCtrl-2.5D over both training-free and training-based methods, attributing this preference to its better trajectory alignment and more natural motion generation.}
    % \vspace{-2mm}
    \label{fig:user_study}
\end{figure}

For a more comprehensive evaluation, we conduct a user study on ObjCtrl-Test, involving fifty participants with experience in AIGC. They were asked to vote for the video results they subjectively found to best align with the given trajectories. The statistical results, shown in Fig.~\ref{fig:user_study}, indicate that approximately 72.95\% of participants preferred ObjCtrl-2.5D over PEEKABOO~\cite{peekaboo} and FreeTraj~\cite{freetraj}, while 63.68\% favored ObjCtrl-2.5D over DragNUWA~\cite{dragnvwa} and DragAnything~\cite{draganything}.

\subsection{Ablation Study}

\noindent\textbf{The Effectiveness of Depth from $\mathbf{I_c}$.}
To evaluate the effectiveness of extending a 2D trajectory to 3D using depth information from the conditional image $\mathbf{I_c}$, we compare the results of ObjCtrl-2.5D conducted on 2D trajectory with depth to results obtained using 3D trajectories in DAVIS~\cite{davis}, where trajectories are extracted from real-world videos with SpatialTracker~\cite{spatracker}. ObjCtrl-2.5D with 3D trajectories achieves an ObjMC score of 92.08, closely matching the 91.42 score obtained by combining a 2D trajectory with depth from $\mathbf{I_c}$. This result indicates that supplementing a 2D trajectory with depth from $\mathbf{I_c}$ can effectively approximate a 3D trajectory, making it valuable for aiding object motion control in T2V generation.

\begin{table}[t]
    \centering
    % \vspace{-0.35cm}
    % \vspace{-0.6cm}
    % \vspace{-1.5mm}
    \caption{\textbf{Quantitative Results of Ablation Study Evaluated on ObjCtrl-Test.} Both scale-wise mask and share warped lattent (SWL) improve object control performance, which SWL outperforms copy-pasting method proposed in FreeTraj~\cite{freetraj}.}
    \vspace{-0.2cm}
    \resizebox{\linewidth}{!}{
    \begin{tabular}{ccc|ccc}
        \toprule
         Scale-wise Mask & SWL          & Copy-pasting & \textbf{FID~$\downarrow$} & \textbf{FVD~$\downarrow$} & \textbf{ObjMC~$\downarrow$}  \\
         \midrule
         \checkmark      &              &              & 240.85 & 27.50 & 122.07 \\
                         & \checkmark   &              & 246.94 & 27.88 & 124.37 \\
         \checkmark      &              &   \checkmark & 248.39 & 28.28 & 138.22 \\
         \checkmark      & \checkmark   &              & 247.48 & 27.82 & 120.37 \\
         \bottomrule
    \end{tabular}
    }
    % \vspace{-0.6cm}
    % \vspace{-12mm}
    \label{tab:ablation}
\end{table}

\begin{figure*}[th]
    \centering
    
    \includegraphics[width=0.99\linewidth]{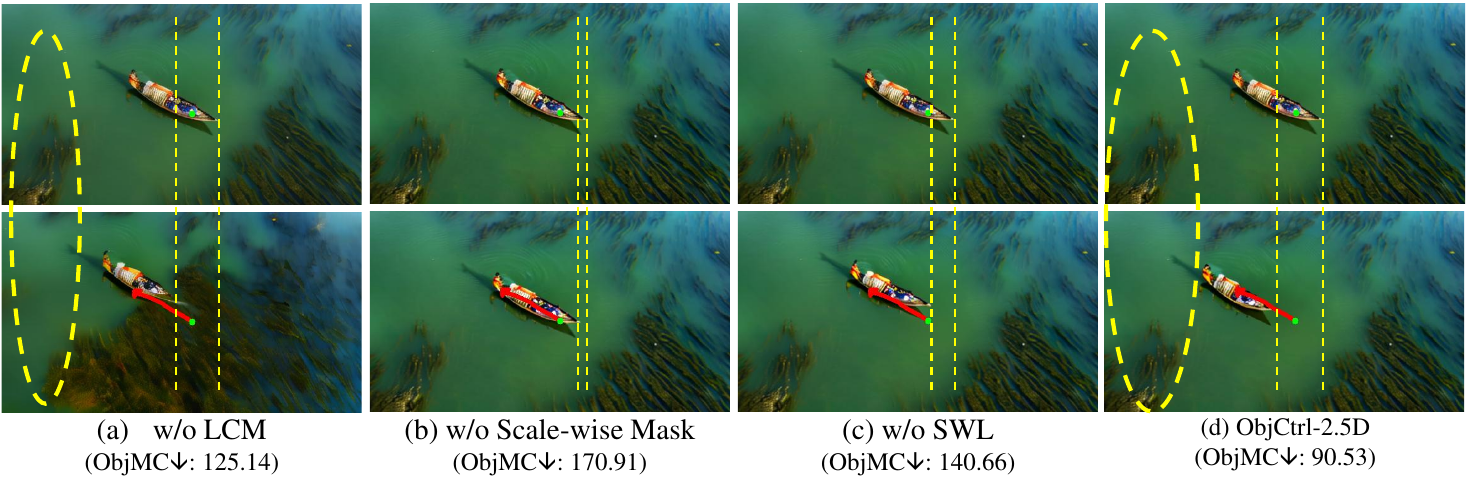}
    \vspace{-0.2cm}
    \caption{\textbf{Qualitative Results of Ablation Studies on LCM, Scale-wise Mask, and SWL.} 
    \revised{(a) Without Layer Control Module (LCM), the original camera control model applies motion control to the entire scene, causing both the object and background to move toward the upper left. (b) Removing the Scale-wise Mask results in an obvious loss of motion control. (c) Removing the Shared Warping Latent (SWL) reduces controllability compared to the full ObjCtrl-2.5D design in (d), as reflected by a higher ObjMC score. (The yellow lines indicate the movement of the boat tail in the scene.)}}
    
    \label{fig:ablation_lcm}
\end{figure*}

\noindent\textbf{The Effectiveness of Layer Control Module and Scale-wise Masks.}
The LCM is designed to adapt the camera motion control module for object motion control by separating the object from the background, enabling independent motion control for each. Without LCM, the base model of ObjCtrl-2.5D typically aligns the trajectory by shifting the entire scene, as shown in Fig.~\ref{fig:ablation_lcm} (a). With LCM, however, the global motion can be segmented into two distinct camera poses for the object and background. Yet, because the features of these two camera poses are spatially merged based on object size, there is a potential risk of losing control over the object's motion. To address this, we introduce scale-wise masks that progressively dilate the merging mask as the feature scale is downsampled.

To assess the effectiveness of the scale-wise mask, we remove the dilation operation and apply the same mask at all scales. This results in an increase in ObjMC score on ObjCtrl-Test from 120.37 to 124.37 (Table~\ref{tab:ablation}). The failed object motion for the boat, as shown in Fig.~\ref{fig:ablation_lcm} (b), highlights this limitation. In contrast, ObjCtrl-2.5D with scale-wise masks successfully drives the target object, as seen in (c) and (d), demonstrating the effectiveness of both the LCM and scale-wise masking.

\begin{figure}[t]
    \centering
    \includegraphics[width=0.99\linewidth]{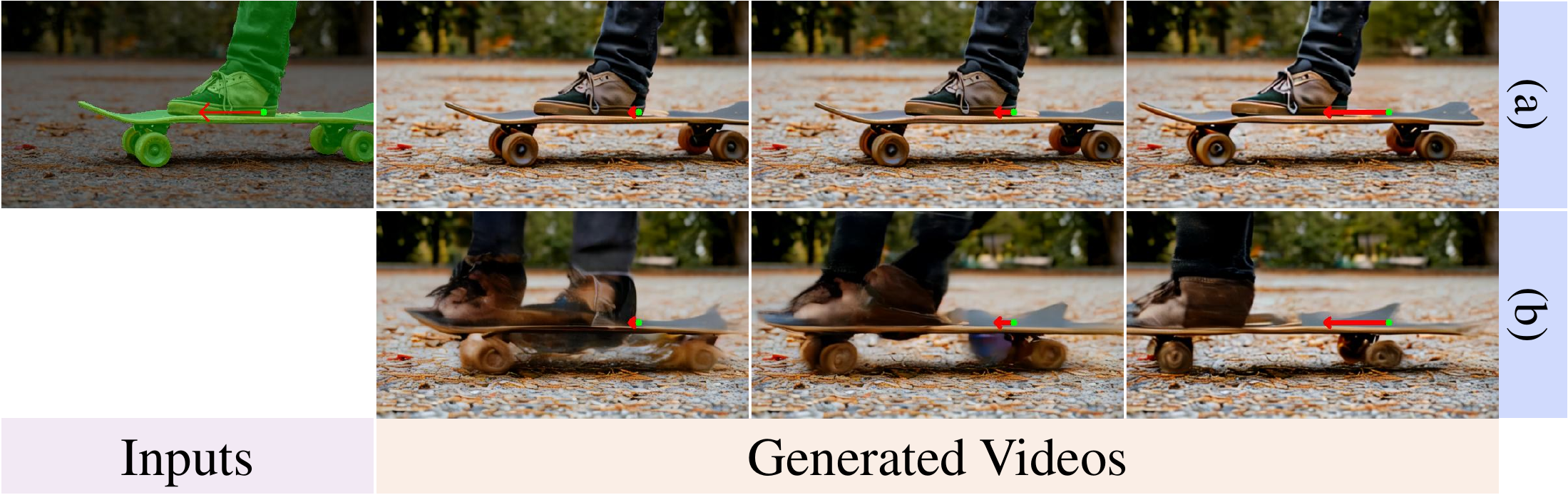}
    % \vspace{-1mm}
    \caption{\textbf{Qualitative Results of Ablation Studies on (a) SWL and (b) Copy-pasting Shared Latent.} The Shared Warping Latent (SWL) in ObjCtrl-2.5D restricts the shared latent specifically within the object's warping areas, effectively avoiding unintended effects on the background while controlling the target object. In contrast, the copy-pasting mechanism used in FreeTraj~\cite{freetraj} coarsely applies the shared latent within bounding boxes, resulting in pronounced artifacts in the generated video.}
    % \vspace{-3mm}
    \label{fig:ablation_swl}
\end{figure}

\noindent\textbf{The Effectiveness of Shared Warping Latent.}
As shown in Fig.~\ref{fig:ablation_lcm} (c) and (d), ObjCtrl-2.5D aligns with the given trajectories more accurately when using SWL compared to settings without it. By sharing latent across frames within the warping object areas, SWL provides strong motion guidance, enhancing trajectory accuracy. Compared to FreeTraj's copy-and-pasting mechanism~\cite{freetraj}, where the shared latent is bounded by a box that includes areas outside the object, SWL achieves a better ObjMC score (120.37 vs. 138.22 in Table~\ref{tab:ablation}) and avoids visible artifacts (Fig.~\ref{fig:ablation_swl}). 
% \textit{However, as with ~\cite{freetraj}, we find that sharing latent across frames can decrease generation quality and is sensitive to sample variations, which are also indicated by the performance drop in terms of FID and FVD in Table~\ref{tab:ablation} compared to the setting without SWL. Given ObjCtrl-2.5D’s robust object motion control with LCM, we recommend using SWL as an enhancement for more challenging cases, ensuring a balance between precise motion control and high-quality video generation.}

\begin{figure}[t]
    \includegraphics[width=0.99\linewidth]{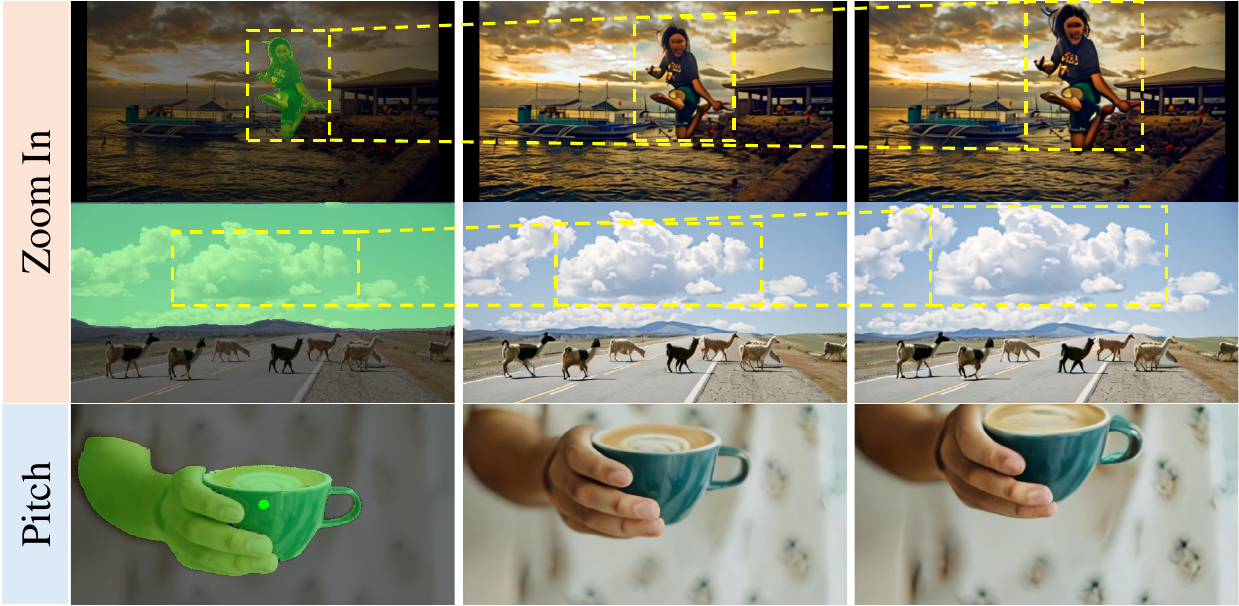}
        % \vspace{-1mm}
        \caption{\textbf{Additional Results with User-Defined Camera Poses.} \revised{ObjCtrl-2.5D allows both the object and background to be manipulated using user-defined camera poses, enabling effects like zooming and pitch, as shown in these examples. More results can be found in the supplementary materials and our \href{https://wzhouxiff.github.io/projects/ObjCtrl-2.5D/}{project page}.}}
        % \vspace{-3mm}
        \label{fig:camera_pose}
\end{figure}

\begin{figure}[th]
    \centering
    \includegraphics[width=0.95\linewidth]{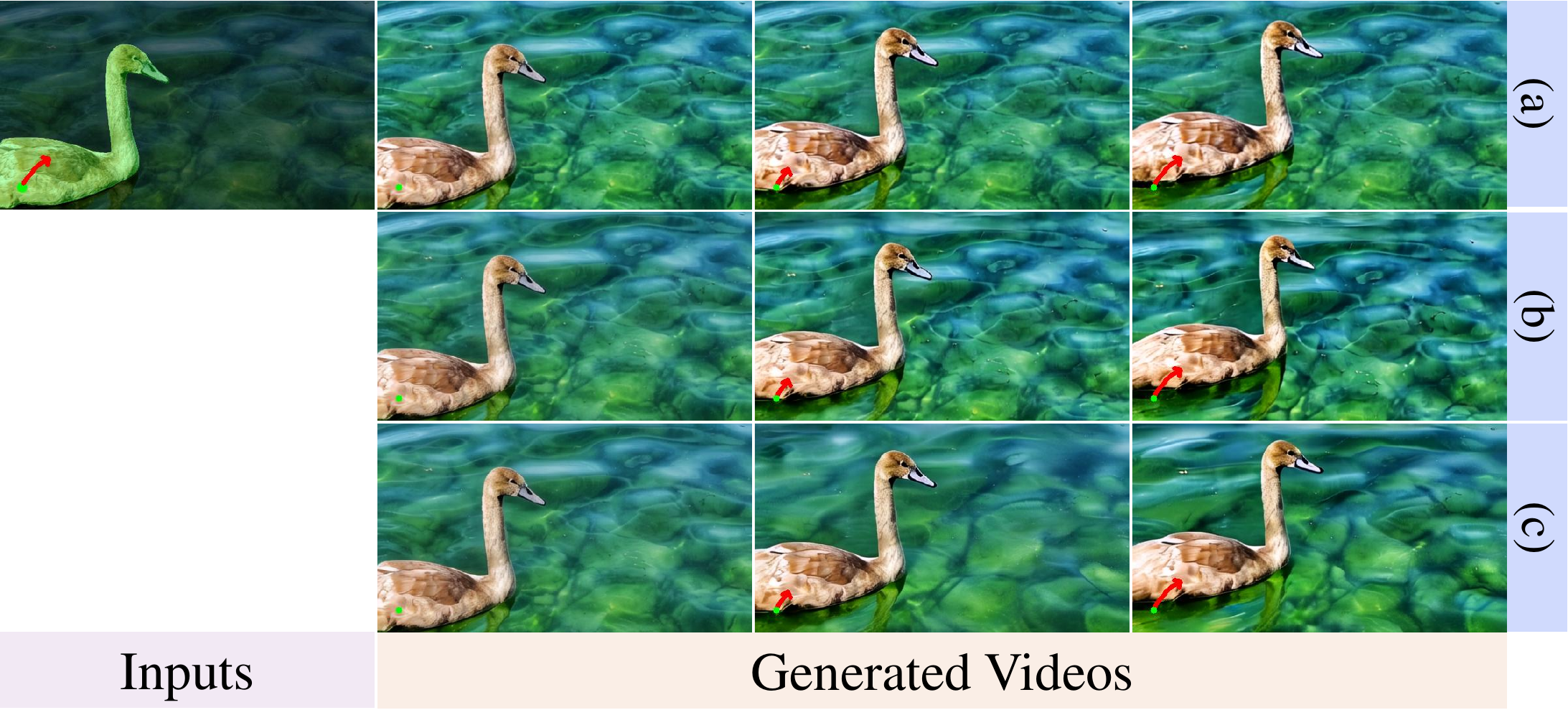}
    % \vspace{-3mm}
    \caption{\textbf{Flexible Background Movement.} (a) Fixed camera poses 
    % ($[\mathbf{I}|\mathbf{0}]$ for all frames) 
    are applied to the background, resulting in the water appearing frozen. (b) A camera movement inversely aligned with the duck's motion is applied to the background, causing the water to move toward the duck. (c) No camera poses are applied to the background, allowing the water to ripple randomly and naturally. In all cases, the object in the generated videos consistently aligns with the given trajectory.
    % (illustrated by the \textcolor{red}{red line with an arrow}). 
    % \textbf{We strongly recommend viewing the demo video for dynamic results.}
    }
    \label{fig:flexible_bg}
\end{figure}

\subsection{More Extensions}
\begin{figure*}[th]
    % \vspace{-2mm}
    \centering
    \begin{tabular}{c}
         \includegraphics[width=0.95\linewidth]{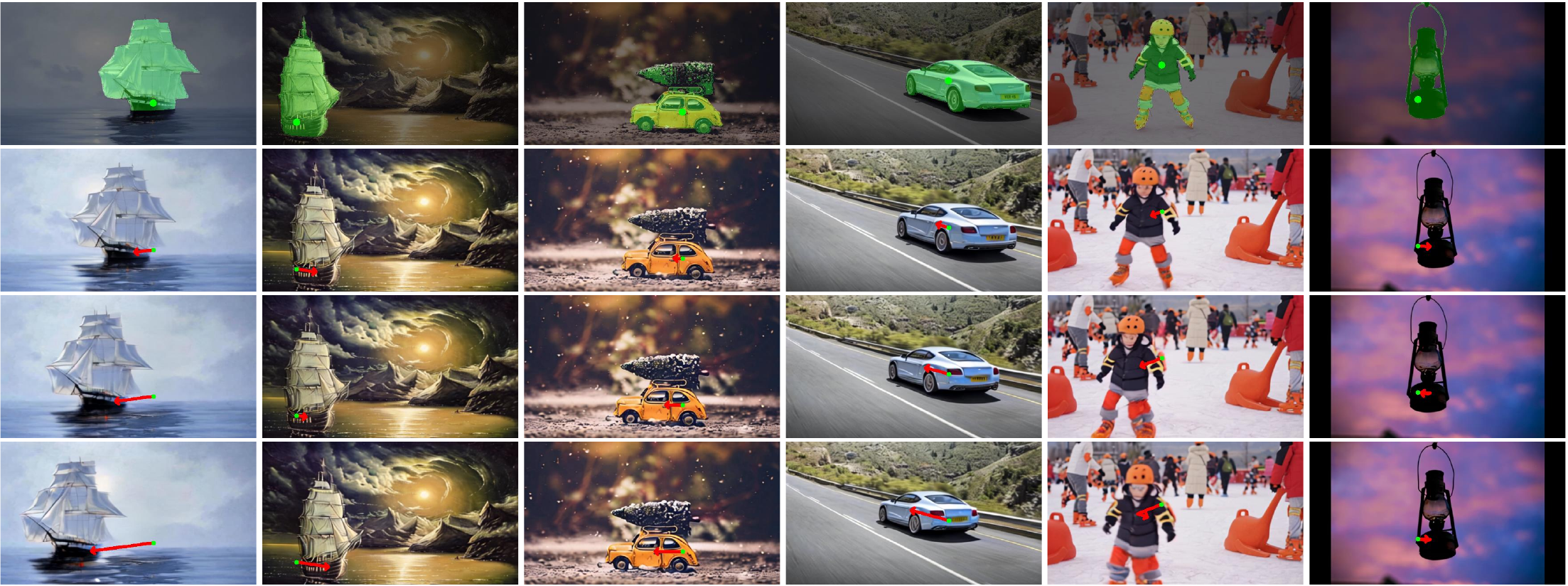} \\
         (a) Guided with Trajectory \\
         \includegraphics[width=0.95\linewidth]{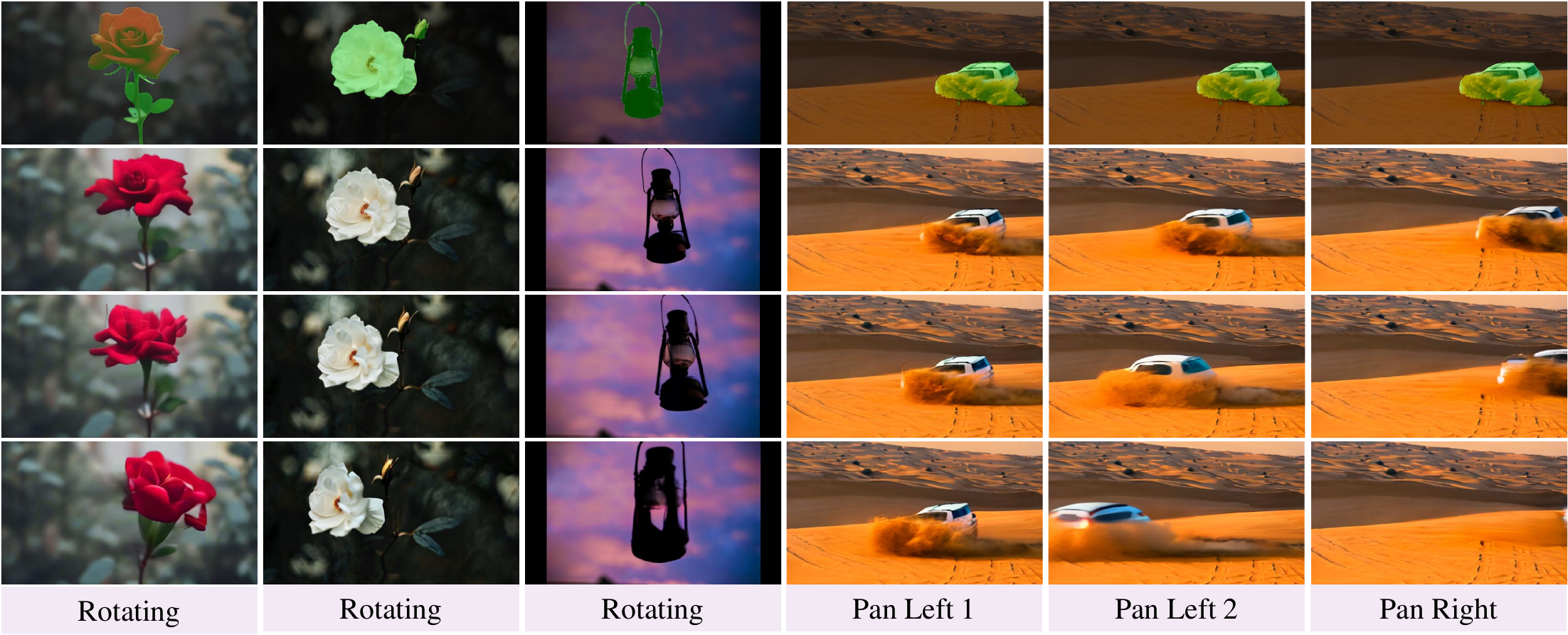} \\
         (b) Guided with Camera Poses Directly
    \end{tabular}
    % \vspace{-2mm}
    \caption{\textbf{More Results of ObjCtrl-2.5D.} ObjCtrl-2.5D supports a wide range of trajectories and camera poses, showcasing its versatility in object motion control. \textbf{We strongly recommend viewing our \href{https://wzhouxiff.github.io/projects/ObjCtrl-2.5D/}{project page} for dynamic results.}}
    \label{fig:more_results}
    % \vspace{-2mm}
\end{figure*}

\noindent\textbf{Control with Customized Camera Poses.}
ObjCtrl-2.5D not only accepts 2D or 3D trajectories as object motion control conditions, but also directly accepts customized camera poses, enabling even more versatile object motion control. As shown in Fig.~\ref{fig:teaser}, given a sequence of anti-clockwise or self-rotating camera poses, ObjCtrl-2.5D can generate videos with spatial rotations (\eg, the snowboarder in the second row) or 3D space rotations (\eg, the rose in the third row). Additionally, more examples, such as zooming in on the object or background and pitching the camera, are provided in Fig.~\ref{fig:camera_pose}. More results are in Fig.~\ref{fig:more_results}.

\noindent\textbf{Flexible Background Movement.}  
The LCM in ObjCtrl-2.5D enables flexible control over background motion by applying different camera poses to background areas. This includes fixed camera poses ($[\mathbf{I}|\mathbf{0}]$) across all frames, poses reversed relative to the object's movement, or no camera poses at all. 
% Detailed visual results can be found in the supplementary materials.
%
As illustrated in Fig.~\ref{fig:flexible_bg}, different camera pose configurations produce diverse background movements. When fixed camera poses ($[\mathbf{I}|\mathbf{0}]$ for all the frames) are applied to the background, the water in (a) is frozen. By introducing a camera movement inversely aligned to the duck's motion, the water in (b) appears to move toward the duck. Furthermore, by removing camera motion control for the background, the water in (c) ripples randomly and naturally. Notably, in all scenarios, the object in the generated videos consistently aligns with the given trajectory.

\begin{figure}[t]
    % \vspace{-4mm}
    \includegraphics[width=0.99\linewidth]{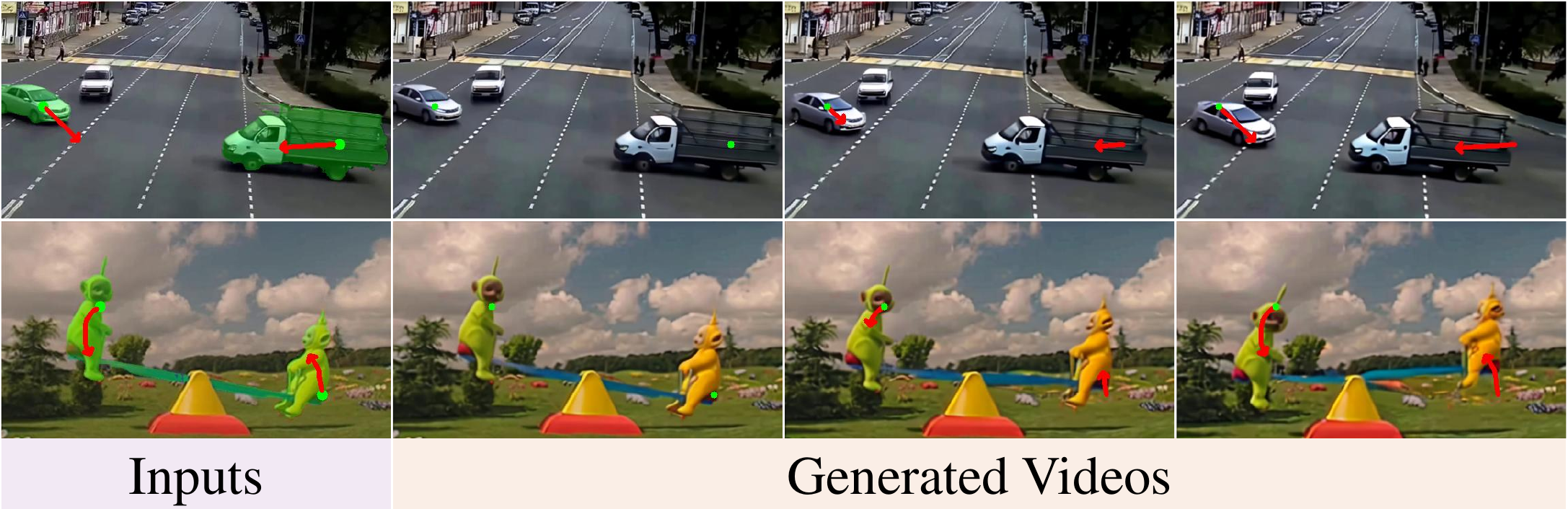}
        % \vspace{-2mm}
        \caption{\textbf{Results with Multiple Trajectories.} \revised{ObjCtrl-2.5D is capable of simultaneously controlling multiple objects.}}
        % \vspace{-3mm}
        \label{fig:multi_traj}
\end{figure}

\noindent\textbf{Multiple Objects.} As illustrated in Fig.~\ref{fig:multi_traj}, ObjCtrl-2.5D is also capable of simultaneously controlling multiple objects. Specifically, the 2D trajectories of different objects can be independently transformed into 3D and further represented as corresponding camera poses. These camera pose features are then applied to their respective target objects using the proposed Layer Control Module (LCM). However, ObjCtrl-2.5D has limitations in handling occluding or overlapping objects, as overlapping masks may confuse the LCM and degrade control precision.

\subsection{Limitations}
\label{sec:limitation}
% \vspace{-2mm}

\begin{figure}[th]
    \centering
    \includegraphics[width=0.99\linewidth]{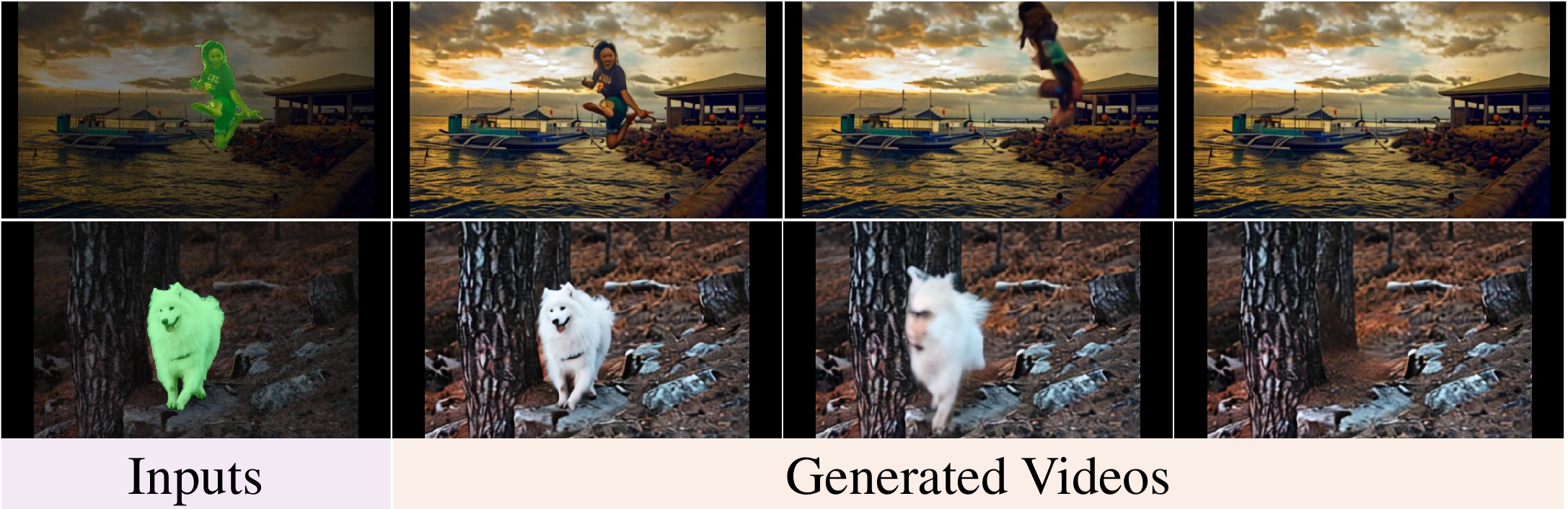}
    \vspace{-2mm}
    \caption{\textbf{Failure Cases.} Due to the limitations of SVD~\cite{svd} in handling large motions, ObjCtrl-2.5D with \textit{high-speed} camera poses results in the object fading out of the scene, leaving only the background. Interestingly, this outcome reveals potential for \textit{image inpainting} applications, as seen in the last frames of the generated videos.}
    \label{fig:failure_cases}
    
\end{figure}

As a training-free method, the quality and motion fidelity of ObjCtrl-2.5D depends on the performance of the underlying video generation model. Since the SVD model struggles with fast-moving objects, ObjCtrl-2.5D is less effective for long trajectories within 14 frames. This limitation can lead to issues such as motion blur, misalignment, or object elimination when handling rapid or complex object movements.
Fig.~\ref{fig:failure_cases} demonstrates how high-speed camera poses can cause the object to fade out of the scene, leaving only the background. Interestingly, this unintended outcome reveals potential for image inpainting applications (see the last frame).
Furthermore, as previously noted, ObjCtrl-2.5D faces challenges when dealing with occluding or overlapping objects, since overlapping masks can interfere with the Layer Control Module. Addressing these complex scenarios is part of our ongoing research.

% \vspace{4mm}

\section{Conclusion}
\label{sec:conclusion}
In this study, we introduce ObjCtrl-2.5D, a novel framework designed to improve object motion control in video generation by incorporating 3D trajectories derived from 2D trajectories and scene depth information. By representing object movement through camera poses, ObjCtrl-2.5D effectively leverages existing Camera Motion Control T2V (CMC-T2V) models to achieve accurate object control without additional training. Our approach includes the development of a Layer Control Module (LCM) to isolate the target object from the background and a Shared Warping Latent (SWL) to enhance control precision by establishing consistent initial object movement. Experimental results demonstrate that ObjCtrl-2.5D largely surpasses existing training-free methods in control accuracy, as validated by both objective and subjective metrics. Additionally, ObjCtrl-2.5D supports complex object movements, such as object rotation, further broadening its application in video generation. This work underscores the value of integrating depth information for realistic video outputs and highlights the potential for future advancements in controllable 3D video generation.
{
    \small
    \bibliographystyle{ieeenat_fullname}
    \bibliography{reference}
}

% WARNING: do not forget to delete the supplementary pages from your submission 
\clearpage
\renewcommand\thesection{\Alph{section}}
\setcounter{section}{0}
\maketitlesupplementary

\noindent The supplementary materials provide additional details and results achieved with the proposed ObjCtrl-2.5D, accompanied by in-depth analyses. \textbf{For a comprehensive understanding, we highly encourage readers to view our \href{https://wzhouxiff.github.io/projects/ObjCtrl-2.5D/}{project page} showcasing dynamic results.} The structure of the supplementary materials is outlined as follows:

\begin{itemize}
    \item Section~\ref{sec:more_2d_to_3d} provides additional details on transforming 2D trajectories into 3D using depth extracted from the conditional image.
    \item Section~\ref{sec:more_extensions} discusses extensions involving customized camera poses. % and flexible background movements.
    \item Section~\ref{sec:more_compared} presents additional comparative results with previous methods.
    % \item Section~\ref{sec:more_results} showcases more results generated using ObjCtrl-2.5D.
    % \item Section~\ref{sec:failure_cases} highlights failure cases and explores the potential of ObjCtrl-2.5D for image inpainting.
\end{itemize}

\section{More Details about 2D Trajectories to 3D}
\label{sec:more_2d_to_3d}

\begin{figure*}[h]
    \centering
    \includegraphics[width=0.95\linewidth]{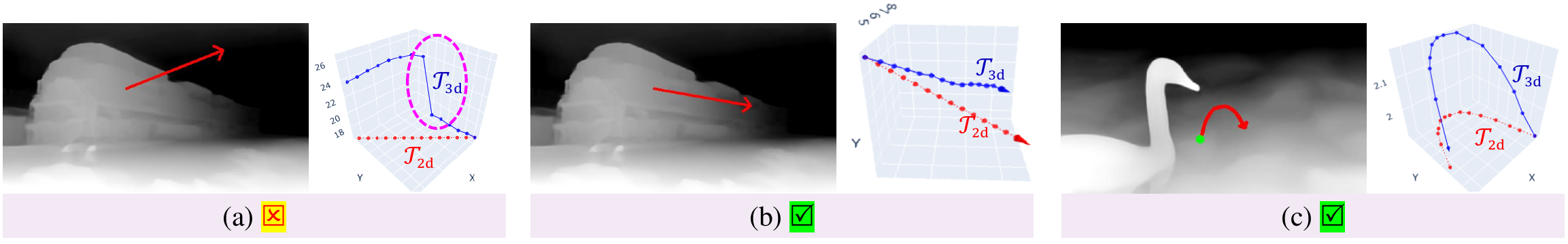}
    \caption{\textbf{Guidelines for Drawing Trajectories.} Drawing 2D trajectories directly on the depth image is recommended, as it ensures smoother depth transitions and avoids abrupt changes (refer to (a)) with the intrinsic depth information. Furthermore, trajectories can be drawn anywhere on the depth image to achieve appropriate depth values without affecting the movement of the target object.}
    \label{fig:2d_to_3d}
\end{figure*}

In this work, ObjCtrl-2.5D extends 2D trajectories to 3D by utilizing depth information, \(\mathbf{D_c}\), extracted from the conditional image \(\mathbf{I_c}\). The depth \(d^i\) of each trajectory point \((x^i, y^i)\) is determined by the corresponding depth value \(\mathbf{D_c}(x^i, y^i)\).  
When the trajectory spans both the foreground object and the background, significant depth variations may occur between consecutive points, as shown in Fig.~\ref{fig:2d_to_3d} (a). This can result in abrupt changes in object movement along the trajectory. To address this, we smooth the 3D trajectory by analyzing its gradient, defined as $grad = d^i - d^{i-1}, i \in [1, N-1]$, and computing the standard deviation of the gradient, $grad_{std}=\mathbf{std}(grad)$. If $grad_{std} > \theta$, the depth \(d^i\) is reset to the initial depth \(d^0\). In this work, we set \(\theta = 0.2\).

To prevent such issues, we recommend drawing the trajectory directly on the depth image, as shown in Fig.~\ref{fig:2d_to_3d} (b) and (c), which inherently provides smoother depth transitions and avoids the abrupt changes shown in (a). Additionally, unlike previous methods such as DragNUWA~\cite{dragnvwa} and DragAnything~\cite{draganything}, which require trajectories to start specifically from the target object, ObjCtrl-2.5D offers greater flexibility. This is because ObjCtrl-2.5D uses object masks to indicate the moving target, while the trajectory serves solely to specify the movement direction. Consequently, trajectories in ObjCtrl-2.5D can be drawn anywhere on the depth image, allowing for the assignment of appropriate depth values at each point (Fig.~\ref{fig:2d_to_3d} (c)).
%
% \textbf{This flexibility is achieved because the trajectory in ObjCtrl-2.5D serves only to indicate object motion and is ultimately transformed into spatially independent camera poses.} Object-specific motion is then implemented using the merged mask introduced by the Layer Control Module in ObjCtrl-2.5D.

\section{More Extensions}
\label{sec:more_extensions}

\begin{figure*}[h]
    \centering
    \includegraphics[width=0.95\linewidth]{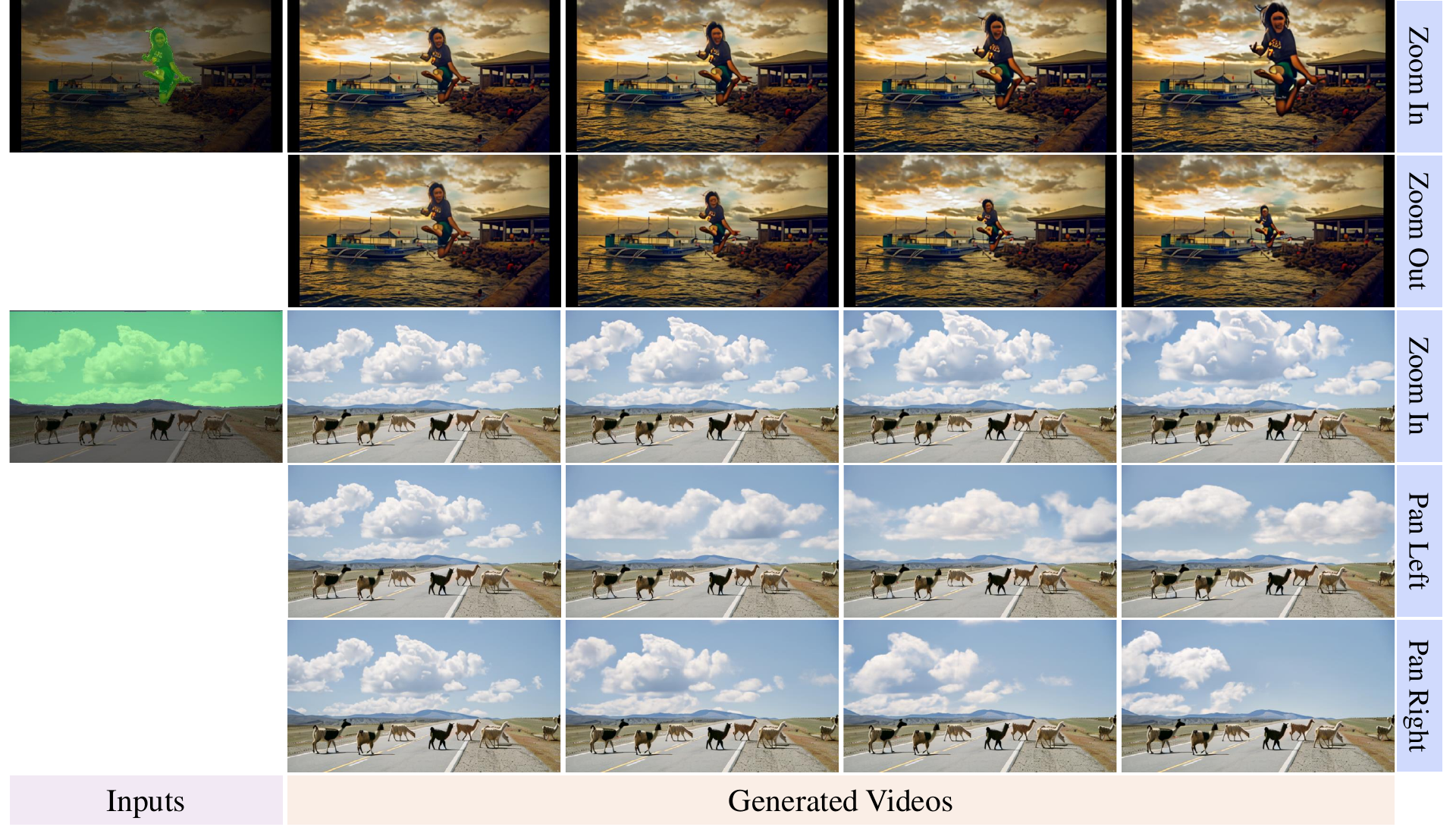}
    \caption{\textbf{Additional Results with User-Defined Camera Poses.} ObjCtrl-2.5D can drive the same sample differently with different camera poses. \textbf{We strongly recommend viewing our \href{https://wzhouxiff.github.io/projects/ObjCtrl-2.5D/}{project page} for dynamic results.}}
    \label{fig:more_camera_poses}
\end{figure*}

\noindent\textbf{Object Control with Customized Camera Poses.}
ObjCtrl-2.5D supports user-defined camera poses for controlling the motion of objects or the background. Beyond the "Zoom In" camera poses presented in the main manuscript, we showcase additional results using various camera poses, including zoom out, pan left, and pan right, as illustrated in Fig.~\ref{fig:more_camera_poses}. The examples demonstrate that ObjCtrl-2.5D can drive the same sample differently with different camera poses, such as the leftward, rightward, and forward movements of the cloud in the second example.

\section{More Compared Results}
\label{sec:more_compared}

\begin{figure*}[h]
    \centering
    \includegraphics[width=0.95\linewidth]{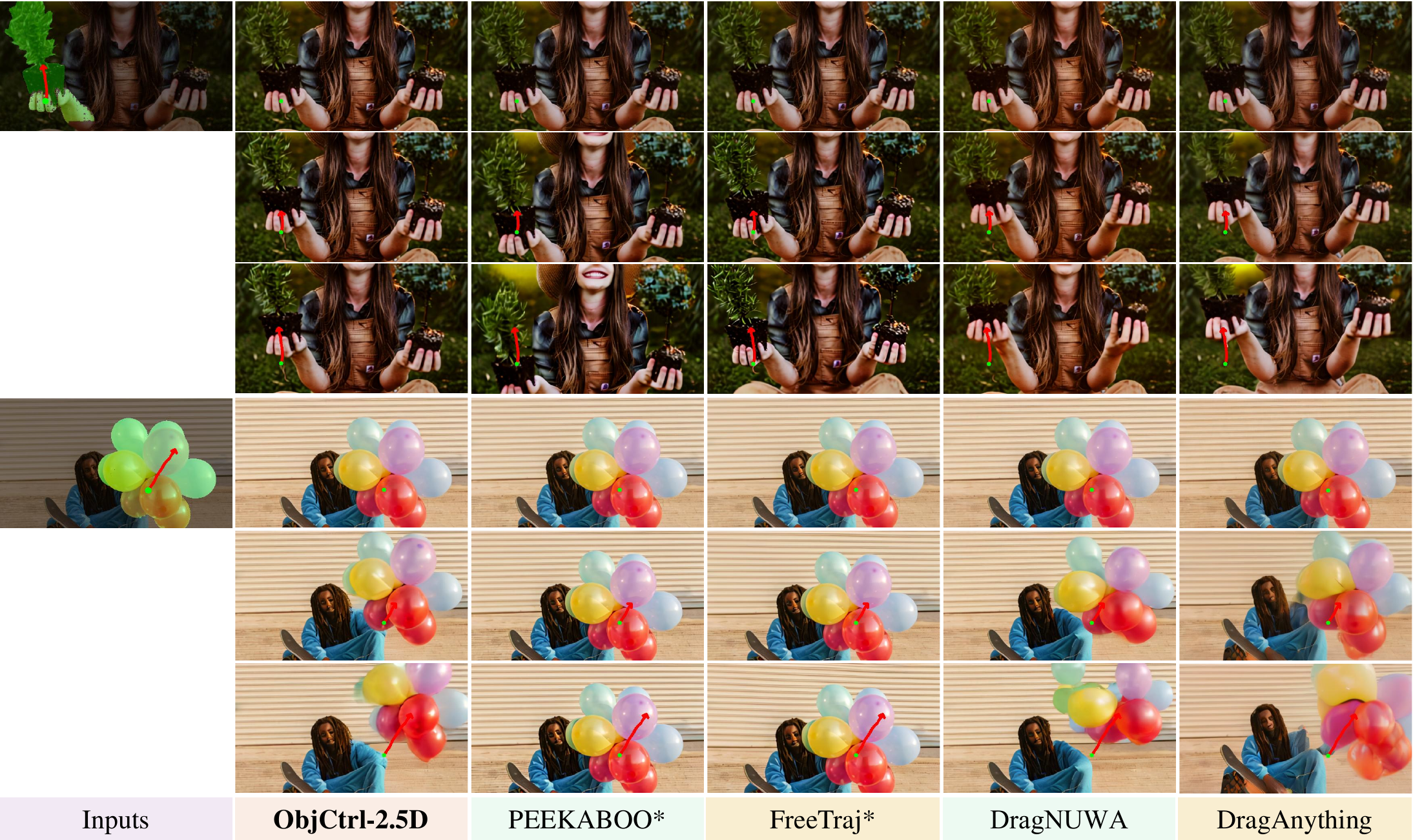}
    \caption{\textbf{More Compared Results with Previous Methods.} ObjCtrl-2.5D outperforms training-free methods (PEEKABOO~\cite{peekaboo} and FreeTraj~\cite{freetraj}) in trajectory alignment and achieves more precise target object movement compared to training-based methods (DragNUWA~\cite{dragnvwa} and DragAnything~\cite{draganything}), which often result in either global scene movement or partial object movement. \textbf{We strongly recommend viewing our \href{https://wzhouxiff.github.io/projects/ObjCtrl-2.5D/}{project page} for dynamic results.}}
    \label{fig:more_comp}
\end{figure*}

We provide additional comparisons with previous methods. As shown in Fig.~\ref{fig:more_comp}, ObjCtrl-2.5D outperforms the training-free methods, including PEEKABOO~\cite{peekaboo} and FreeTraj~\cite{freetraj}, in trajectory alignment. While training-based methods like DragNUWA~\cite{dragnvwa} and DragAnything~\cite{draganything} also achieve good trajectory alignment, they often rely on global movement or parts of the object movement rather than targeting the specific object. In contrast, ObjCtrl-2.5D incorporates a Layer Control Module, enabling relatively precise control over the specific object with minimal impact on other areas of the scene, while maintaining natural video generation. 
% \textbf{We strongly recommend viewing the demo video for dynamic results.}

\end{document}